\documentclass[journal]{IEEEtran}

\usepackage{graphicx}
\usepackage{amsmath}
\usepackage{amssymb}
\usepackage{booktabs}
\usepackage{array}
\usepackage{algorithm}
\usepackage{algpseudocode}
\usepackage{xcolor}
\usepackage{multirow}
\usepackage{balance}
\usepackage{pifont}
\usepackage{eurosym}
\usepackage{threeparttable}
\usepackage{placeins}
\usepackage[colorlinks=true, linkcolor=black, citecolor=black, urlcolor=blue]{hyperref}

\begin{document}

\title{
Staged Multi-Agent Training for Human–Exoskeleton Co-Adaptation: Simulation-Based Learning and Real-World Metabolic and Biomechanical Validation
}

\author{Yifei~Yuan,
        Jakob~Wolf,
        Ghaith~Androwis,
        and~Xianlian~Zhou%
\thanks{Yifei Yuan, Jakob Wolf and Xianlian Zhou are with the Department of Biomedical
Engineering, New Jersey Institute of Technology, Newark, NJ 07102, USA
(e-mail: yy72@njit.edu; jw768@njit.edu; alexzhou@njit.edu).}%
\thanks{Ghaith Androwis is with Kessler Foundation, West Orange, NJ 07052, USA
(e-mail: GAndrowis@kesslerfoundation.org).}%
\thanks{This work was partially supported by the National Institute on Disability, Independent Living, and Rehabilitation Research (NIDILRR) funded Rehabilitation Engineering Research Center Grant 90REGE0025-01-00 and NSF Award \#2524089. The human study was approved by the Institutional Review Board under Protocol No.\ 2305033091R003.}%
\thanks{This work has been submitted to the IEEE for possible publication. Copyright may be transferred without notice, after which this version may no longer be accessible.}}

\markboth{}{}

\maketitle

\begin{abstract}
Learning-based controllers can enable exoskeleton assistance to be trained entirely in physics-based simulation, but achieving effective assistance on real users remains challenging because the human and device continuously adapt to one another. We investigate a Staged Multi-Agent Training (SMAT) approach that jointly trains a musculoskeletal human actor and a bilateral hip-exoskeleton actor (control policy) to learn coordinated assistance strategies while accounting for human--device co-adaptation. The resulting policy was transferred to a physical hip exoskeleton and evaluated in eight healthy adults across users and locomotion conditions without subject-specific retraining. Indirect calorimetry was used to compare no-exoskeleton, passive, and active conditions. Active assistance reduced net metabolic rate by 19.7\% relative to the passive device ($p<0.001$). The policy produced predominantly positive hip mechanical power across subjects, with a positive-power ratio of 0.984, and maintained effective assistance across walking speeds and terrains beyond the training conditions. These results demonstrate that simulation-based multi-agent co-adaptation can produce a single transferable exoskeleton policy that elicits meaningful physiological benefits in real users without subject-specific retraining, providing a framework for reducing the experimental burden of human-in-the-loop controller development.
\end{abstract}

\begin{IEEEkeywords}
Exoskeletons, physical human--robot interaction, reinforcement learning, co-adaptation, metabolic cost.
\end{IEEEkeywords}

\section{Introduction}
\IEEEPARstart{L}{ower-limb} exoskeletons offer a promising route toward gait rehabilitation and physical augmentation, but their usefulness hinges on delivering assistance that fits the way each individual walks~\cite{rodriguez2021systematic,baud2021review}. Consider what happens when an exoskeleton begins to assist: as the device applies torque at the joint, it alters the mechanical dynamics of walking, and the wearer's neuromuscular system does not simply absorb this change but actively responds to it, reorganizing muscle recruitment and inter-limb coordination to accommodate the modified dynamics~\cite{poggensee2021adaptation}. The controller and wearer thereby form a continuously interacting loop, each reshaping the conditions under which the other operates. For the controller, this makes assistance a non-stationary optimization problem, with its target shifting as the wearer keeps adapting. Automation of this controller development or tuning process is therefore difficult because the controller must be synthesized against a moving target rather than a fixed plant.

A growing body of work now learns exoskeleton assistance controllers directly in physics-based musculoskeletal simulation, avoiding the lengthy human-in-the-loop tuning that handcrafted controllers require. The human-in-the-loop optimization process personalizes assistance directly on the wearer but requires substantial per-user experimental time and specialized instrumentation \cite{zhang2017human,ding2018human,slade2024human}, and must be repeated for each user and condition. Simulation-based learning instead moves this optimization entirely into simulation, so that a single controller is synthesized once and deployed across users and conditions without per-user experiments. Early learning frameworks coupled a neural control policy to a musculoskeletal model to learn robust assistance in simulation~\cite{luo2023robust}, and a policy trained entirely in simulation was then shown to transfer to hardware and reduce metabolic cost without subject-specific experiments~\cite{luo2024experiment}. Building on this paradigm, later work has extended learning-in-simulation in several directions: muscle-synergy priors with policy distillation transfer speed- and slope-aware hip assistance to hardware~\cite{park2026learning}; reflex-based musculoskeletal models are paired with reinforcement learning for end-to-end hip-exoskeleton policies that lower metabolic cost in real users~\cite{barati2026end}; human-aligned simulation optimizes hip assistance parameters without real-world experiments~\cite{leem2026exo}; and the approach extends to asymmetric gait~\cite{yuan2026gait} and multi-joint tasks such as squatting~\cite{ratnakumar2026reinforcement,ratnakumar2025optimizing}, sit-to-stand~\cite{ratnakumar2026predicting}, and running~\cite{simos2025reinforcement}. Whole-body metabolic validation on real users, the standard benchmark for assistive walking, has so far been reported for only a small number of these controllers~\cite{luo2024experiment,barati2026end}. Across these advances, the mutual adaptation between human and device is seldom treated as an explicit learning problem, and the resulting non-stationarity is managed implicitly rather than addressed directly.

Optimizing the human and exoskeleton policies simultaneously can be unstable because each policy continually changes the environment in which the other is adapting. This interaction can lead to oscillatory torques, suboptimal or poorly timed assistance, or even training divergence. To address this challenge, we previously introduced Staged Multi-Agent Training (SMAT)~\cite{yuan2026smat}, a multi-stage curriculum for human--exoskeleton co-adaptation. SMAT decomposes the policy-training problem into a curriculum that sequentially isolates gait acquisition, mass adaptation, assistance-timing pre-training, and full co-adaptation, allowing each challenge to be addressed independently before the two policies are jointly optimized. Such decomposing can improve training stability~\cite{bengio2009curriculum} and has proven effective for learning physiologically plausible control in high-dimensional musculoskeletal systems~\cite{chiappa2024acquiring}. More importantly, this staged strategy is consistent with the process of human adaptation, in which a wearer first establishes a stable gait before learning to integrate active assistance into their locomotor strategy.

In that work~\cite{yuan2026smat}, the learned SMAT controller reduced simulated activation of the right-side hip muscles by 10.1\,\% and, once deployed to hardware, delivered consistent positive mechanical power across five participants without per-subject retraining. These results, however, leave the crucial question open: \emph{Can a co-adaptive controller synthesized entirely in simulation, without any per-user tuning, translate its simulated muscle offloading into a measurable reduction in the metabolic cost of walking for real users?} Such a translation is not guaranteed. A wearer may redistribute effort across muscle groups rather than reduce overall expenditure, and the added mass and rigid structural constraints of a powered hip exoskeleton impose a metabolic penalty before any assistance is rendered. Active assistance must overcome this penalty before it can yield a net benefit over unassisted walking. Answering the question therefore requires measuring whole-body net metabolic rate, the standard benchmark for assistive devices~\cite{seo2016fully,slade2024human}.

This work extends Staged Multi-Agent Training (SMAT), a simulation-based framework for human–exoskeleton co-adaptation, from physics-based training to real-world deployment and physiological validation. A single final policy trained entirely in simulation was deployed on a hip exoskeleton and evaluated in eight healthy adults without subject-specific retraining. The principal contributions are as follows:

\begin{enumerate}
\item \textbf{Simulation-trained co-adaptive assistance with real-world physiological validation.} Across eight subjects, SMAT assistance reduced net metabolic rate by 19.7\,\% relative to the passive device ($p<0.001$) and by 9.1\,\% relative to the no-exoskeleton condition ($p<0.01$). Although the 5.94 kg exoskeleton increased metabolic rate by 13.1\,\% in the passive condition relative to no exoskeleton, the simulation-trained policy more than compensated for this device penalty. Evaluating all three conditions within the same subjects isolates the device penalty from the assistance benefit (assistance vs. passive);
\item \textbf{Biomechanical validation of the learned assistance strategy.} Joint-level analysis across the no-exoskeleton, passive, and SMAT conditions demonstrates that the deployed policy produces predominantly positive hip mechanical power with appropriate timing in real users, yielding a positive-power ratio of 0.984 across subjects. The assistance also markedly increased hip range of motion, providing evidence of altered human--exoskeleton coordination consistent with co-adaptation.
\item \textbf{Generalization of a single simulation-trained policy across users and locomotion conditions.} Without subject-specific retraining or speed- or terrain-specific policy switching, the same deployed policy maintained predominantly positive assistance ($R_\mathrm{pos}\geq0.96$) across walking speeds from 0.76 to 1.74 m/s and across outdoor terrains including ramps and stairs. 
\end{enumerate}

\section{Staged Multi-Agent Training}

This section summarizes and extends the SMAT learning-in-simulation formulation first introduced in our earlier conference paper~\cite{yuan2026smat}, covering the multi-agent framework and the four-stage curriculum used to obtain the final policy that was deployed to the exoskeleton hardware.

\subsection{Multi-Agent Reinforcement Learning Framework}
\label{subsec:framework}
\begin{figure*}[t]
    \centering
    \includegraphics[width=\textwidth]{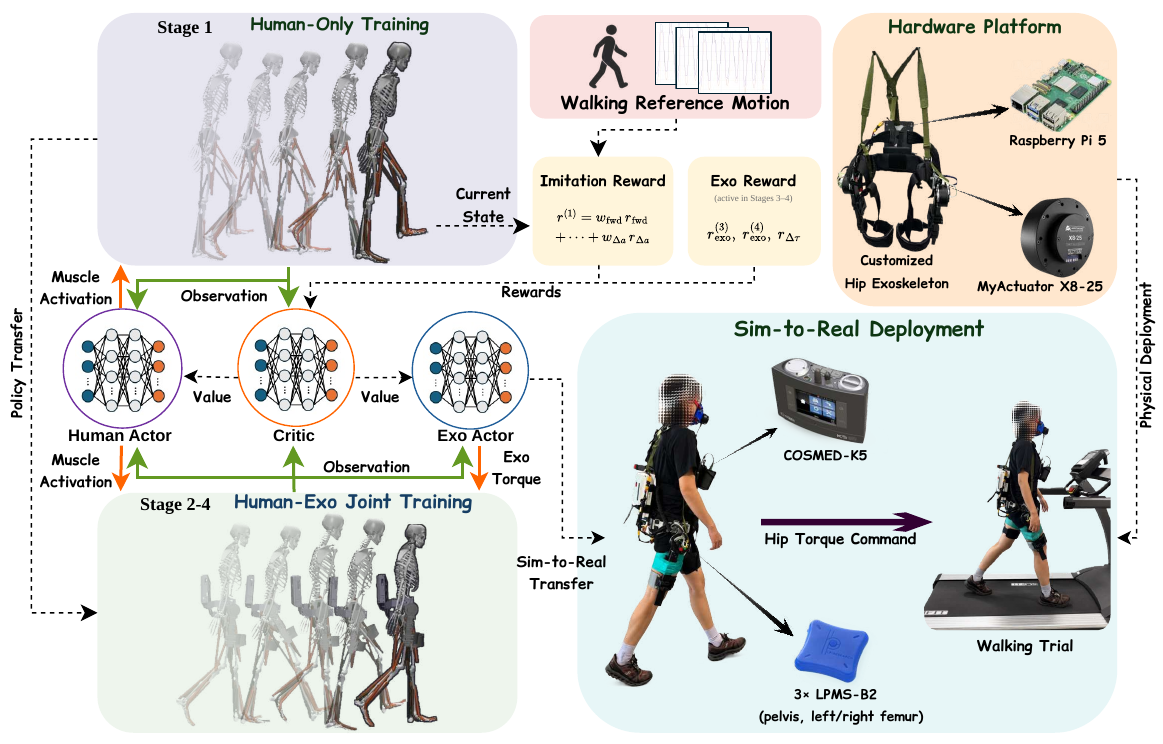}
    \caption{Overview of the SMAT framework. Two PPO-based actors, a musculoskeletal human actor $\pi_h$ and an exoskeleton actor $\pi_e$, interact with a shared physics simulation through a shared critic $V_\psi$. Stage~1 trains $\pi_h$ alone. Stage~2 adapts $\pi_h$ with added exoskeleton mass. Stage~3 freezes $\pi_h$ and trains $\pi_e$ for positive assistance timing. Stage~4 co-adapts both actors. The imitation reward is active throughout, whereas the exoskeleton reward applies only after the device is introduced. The Stage~4 policy is deployed via sim-to-real transfer on a Raspberry~Pi~5 driving bilateral MyActuator X8-25 actuators.}
    \label{fig:framework}
\end{figure*}

We cast hip exoskeleton assistance as a multi-agent reinforcement learning (MARL) problem~\cite{luo2024experiment,tan2025myoassist,lowe2017multi} and realize it within a configurable actor--critic framework (Fig.~\ref{fig:framework}). Two agents learn concurrently in a shared physics simulation coupling a musculoskeletal human model to a bilateral hip exoskeleton: a human actor $\pi_h$ governing muscle control and an exoskeleton actor $\pi_e$ commanding assistive torque.

\looseness=-1
The human actor maps a musculoskeletal body-state observation to 26 muscle excitations $\mathbf{e}\in[0,1]^{26}$ (13 per side) through an MLP with hidden sizes $[256,128]$. Fig.~\ref{fig:hardware} shows the musculoskeletal model with the bilateral muscles, wearing the designed exoskeleton. 
The exoskeleton actor receives an 18-dimensional observation $\mathbf{o}_e$ comprising a three-step history of bilateral hip flexion angles and angular velocities (12 dimensions) together with a three-step history of its own normalized torque outputs $[\hat{u}_r,\hat{u}_l]$ (6 dimensions), with no explicit gait-phase or timing variable provided. It outputs normalized bilateral hip torque commands $\hat{\mathbf{u}}\in[-1,1]^{2}$, scaled by stage-dependent torque limits to produce physical assistance, through an MLP with hidden sizes $[128,64]$. Both actors share a common critic with hidden sizes $[256,128]$ (Fig.~\ref{fig:framework}).

\subsection{Multi-stage Curriculum Training}
\label{subsec:curriculum}

To facilitate co-adaptive learning, we split training into four sequential stages, each isolating a single adaptation challenge. The human actor first learns baseline gait, then adapts to the added device mass and inertia. With that gait stabilized, the exoskeleton actor learns assistance timing against the frozen human policy, and finally both actors co-adapt under full torque. This staged progression improves training stability and convergence. Algorithm~S1 of the supplementary material summarizes the schedule, the policy freezing, and the reward activation across stages.

\textbf{Stage 1: Human Baseline Gait Learning.}
Stage~1 trains the human actor alone, before the exoskeleton is introduced. The result is a stable gait that tracks a reference trajectory and serves as the starting point for the later stages. The per-timestep reward $r^{(1)}$ sums a set of weighted terms, with the imitation terms adopted from~\cite{tan2025myoassist}:
\begin{equation}
\begin{split}
r^{(1)} = \;&w_{\text{fwd}}\,r_{\text{fwd}}
           + w_{\text{qpos}}\,r_{\text{qpos}}
           + w_{\text{qvel}}\,r_{\text{qvel}}\\
           &+ w_{\text{mus}}\,r_{\text{muscle}}
           + w_{\Delta a}\,r_{\Delta a}.
\end{split}
\label{eq:reward_stage1}
\end{equation}

The first three terms drive gait imitation, tracking the target forward speed and the reference joint trajectories:
\begin{subequations}
\label{eq:reward_imitation}
\begin{align}
r_{\text{fwd}}  &= \Delta t \cdot
  \exp\!\left(-5\,(v - v^*)^2\right),
  \label{eq:rfwd}\\
r_{\text{qpos}} &= \Delta t
  \sum_{i\in\mathcal{J}} w_i\,
  \exp\!\left(-8\,(q_i - q_i^{\text{ref}})^2\right),
  \label{eq:rqpos}\\
r_{\text{qvel}} &= \Delta t
  \sum_{i\in\mathcal{J}} w_i\,
  \exp\!\left(-8\,(\dot{q}_i - \rho\,\dot{q}_i^{\text{ref}})^2\right).
  \label{eq:rqvel}
\end{align}
\end{subequations}
The remaining two terms regulate muscle activity, penalizing the overall activation level and its rate of change:
\begin{subequations}
\label{eq:reward_muscle}
\begin{align}
r_{\text{muscle}} &= -\frac{\Delta t}{N_m}
  \sum_{j=1}^{N_m} a_j,
  \label{eq:rmuscle}\\
r_{\Delta a}      &= \frac{\Delta t}{N_m}
  \sum_{j=1}^{N_m}
  \exp\!\left(-4\,(a_{j,t}-a_{j,t-1})^2\right).
  \label{eq:rdeltaa}
\end{align}
\end{subequations}

Here $v$ is the forward pelvis speed, $v^*{=}1.25$\,m/s is the target speed, $N_m{=}26$ is the total number of muscles, and $a_{j,t}\!\in\![0,1]$ is the activation of muscle $j$ at timestep $t$ with $a_{j,t-1}$ the previous-step activation. $\mathcal{J}$ is the set of tracked joints with per-joint imitation weights $w_i$ (listed in Table~S3 of the supplementary material), and $\rho = v^*/\dot{q}_{\text{pelvis}}^{\text{ref}}(t)$ scales the reference joint velocities to the target walking speed. The control step is $\Delta t{=}0.02$\,s, and multiplying each term by $\Delta t$ makes the reward magnitude independent of control frequency.

\begin{figure}[t]
\centering
\includegraphics[width=0.9\columnwidth]{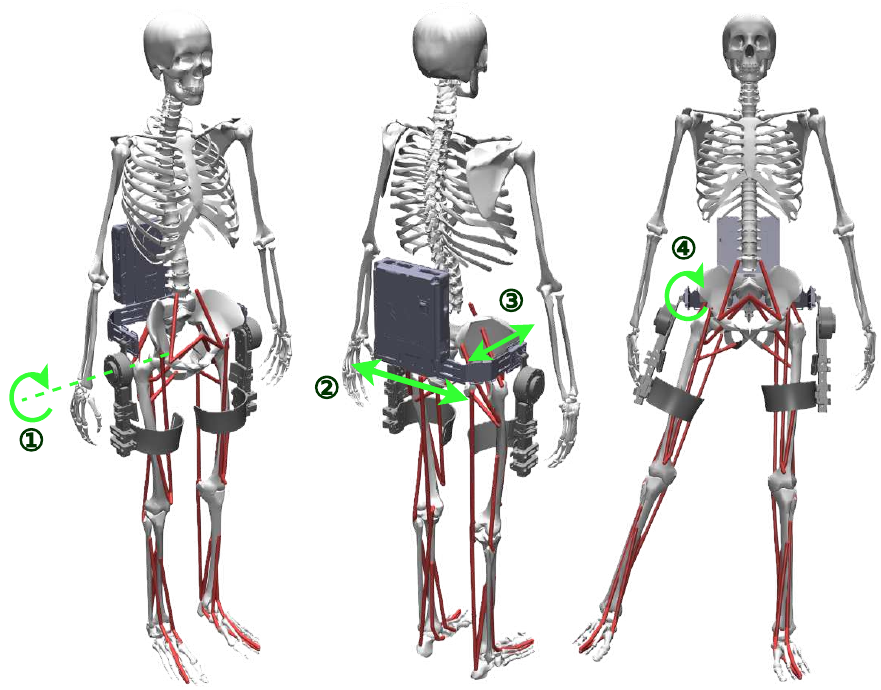}
\caption{Bilateral hip exoskeleton from three views, shown on the 26-muscle musculoskeletal model used for training (13 muscles per side, red). \ding{192}~Actuated flexion/extension joint; the dashed line marks the rotation axis. \ding{193}~Waist width adjusts to the pelvis. \ding{194}~Lateral extensions slide anteroposteriorly to align the actuator with the hip joint center. \ding{195}~Passive abduction/adduction hinge.}
\label{fig:hardware}
\end{figure}

\textbf{Stage 2: Adaptation to Added Exoskeleton Mass.}
Stage~2 continues to train the human policy with the attached exoskeleton providing no assistance to the musculoskeletal model. The pelvis-mounted exoskeleton frame and bilateral thigh links are parented to the corresponding body segments, so the device mass and inertia enter the body dynamics. The exoskeleton actuators apply torque directly to the human hip flexion joints, and the passive abduction joints (Fig.~\ref{fig:hardware}) provide lateral compliance. 
The human policy trains under the same Stage~1 reward, Eqs.~\eqref{eq:reward_stage1}, \eqref{eq:reward_imitation}, and~\eqref{eq:reward_muscle}, and adapts its gait to the added mass before any active assistance begins.

\textbf{Stage 3: Learning of Assistance Timing with Frozen Human Policy.}
Stage~3 freezes the human actor and trains the exoskeleton policy alone, treating the stabilized Stage~2 walking pattern as a fixed environment in which assistance timing can first be learned. The hip imitation terms are disabled here, preventing the exoskeleton from exploiting the tracking rewards by resisting hip motion instead of genuinely assisting the hip flexors and extensors.

A hip muscle activation term is introduced in this stage:
\begin{equation}
r_{\text{hip-act}}
=
-\Delta t \cdot \frac{1}{|\mathcal{M}_{\text{hip}}|}
\sum_{m \in \mathcal{M}_{\text{hip}}}
a_m^2,
\label{eq:rhipact}
\end{equation}
where $a_m \in [0,1]$ is the activation of muscle $m$, and $\mathcal{M}_{\text{hip}}$ comprises major hip flexors and extensors such as the iliopsoas, gluteus maximus, rectus femoris, and hamstrings on both legs. To encourage positive assistance, we also apply a reward based on the sign of the delivered power and on the magnitude of the assistance torque:
\begin{equation}
r_{\text{exo}}^{(3)}
=
\Delta t \sum_{j\in\{l,r\}}
\alpha_d\,\hat{u}_j^2\,
\mathrm{sign}(\hat{u}_j \omega_j),
\label{eq:rexostage3}
\end{equation}
where $\hat{u}_j\in[-1,1]$ is the normalized torque command from the exoskeleton actor and $\omega_j$ the hip joint angular velocity, with $\alpha_d = 0.5$. In this stage, the exoskeleton torque limit is kept at $\tau_{\max}=6$\,Nm to limit its perturbation of human control and focus the exploration on learning beneficial assistance patterns driven by the two new rewards. 

\textbf{Stage 4: Human--Exoskeleton Co-Adaptation.}
The final stage co-adapts both actors together. Stage~4 loads the Stage~3 exoskeleton policy, unfreezes the human actor, and raises the torque limit to the full motor capacity $\tau_{\max}=25$\,Nm (Algorithm~S1). The human actor's observation is augmented with the current normalized exoskeleton torques $\hat{u}_j$ in two dimensions (left and right). We retain the Stage~2 weights on the original dimensions and randomly initialize weights associated with the two new dimensions. This preserves the learned walking policy while allowing the human actor to develop responses to exoskeleton torque during co-adaptation.

Both actors are now trained jointly, sharing a single critic over the full system state. Because the Stage~3 reward $r_{\text{exo}}^{(3)}$ tends to drive the exoskeleton torque into saturation, Stage~4 replaces it with a power- and smoothness-based reward $r_{\text{exo}}^{(4)}$:
\begin{equation}
r_{\text{exo}}^{(4)}
=
\Delta t
\sum_{j \in \{l,r\}}
\left(
\alpha \hat{u}_j \hat{\omega}_j
-
\beta \hat{u}_j^2
-
\lambda_s \max\!\left(0, |\hat{u}_j|-\delta\right)^2
\right),
\label{eq:rexoreward_stage4}
\end{equation}
where $\hat{\omega}_j=\mathrm{clip}(\omega_j/\omega_s,-1,1)$ is the normalized joint angular velocity (with $\omega_s=2.0~\mathrm{rad/s}$). The first term reflects mechanical power, rewarding torque that assists joint motion and penalizing torque that opposes it. The second term penalizes torque magnitude to discourage unnecessarily large torques. The third term imposes a penalty beyond the threshold $\delta$, suppressing actuator saturation when the other terms push the torque upward. We set $\alpha=0.3$, $\beta=0.15$, $\lambda_s=2.0$, and $\delta=0.8$. To discourage abrupt torque change for smooth assistance, a torque-rate penalty is additionally applied:
\begin{equation}
r_{\Delta\tau}
=
-\Delta t
\sum_{j \in \{l,r\}}
\left(\hat{u}_{j,t}-\hat{u}_{j,t-1}\right)^2,
\label{eq:ractionrate}
\end{equation}
where subscripts ${t}$ and ${t-1}$ indicate the current and previous time steps.

Two auxiliary stability penalties are also included in Stages~3 and~4. The joint constraint force penalty $r_{\text{constraint}} = -\Delta t \cdot \max_k(|f_k^{\text{joint}}|/mg)$ penalizes joint limit forces normalized by body weight $mg$, and the foot contact force penalty $r_{\text{foot}} = -\Delta t \cdot \max(0,\,(|f_r|+|f_l|)/mg - 1.2)$ penalizes total foot contact forces above $1.2$ times body weight. Both penalties are inactive in Stages~1 and~2, where the imitation rewards alone maintain stable gait. Stage~4 produces the final co-adapted policies, and Table~S2 of the supplementary material lists the reward weights for all stages.

The curriculum hyperparameters are listed in Table~S4 of the supplementary material. All training ran on an Intel Xeon W-2145 (3.70\,GHz, 8-core), taking approximately 28\,h per 100\,M simulation steps.

\section{Human Experiment}
\label{sec:human}
This section describes how the exoskeleton policy obtained in Stage~4 was transferred to hardware and evaluated in human walking, completing the simulation-to-hardware pipeline outlined in Fig.~\ref{fig:framework}.

\begin{figure*}[t]
\centering
\includegraphics[width=\textwidth]{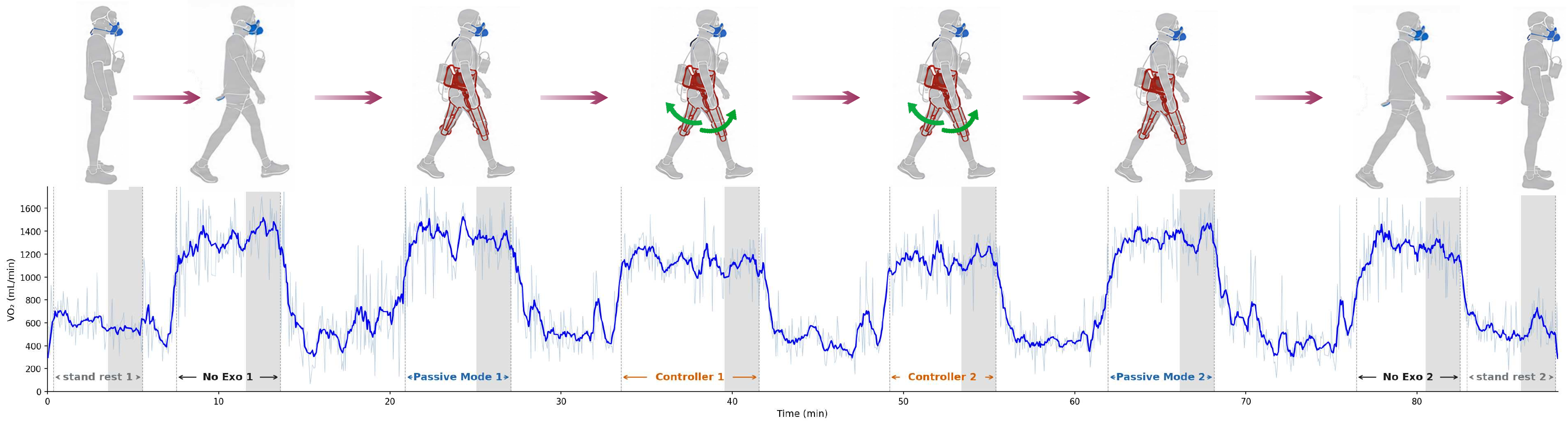}
\caption{Experimental protocol and a representative recording. \emph{Top:} the symmetric (order-balanced) condition sequence (standing rest, no-exo, passive, the two active controllers in randomized order, then passive, no-exo, and rest repeated), illustrated by the subject in each condition (green highlights the active exoskeleton trials). \emph{Bottom:} continuous oxygen uptake ($\dot V_{O_2}$, mL/min); shaded bands mark the steady-state windows used for averaging.}
\label{fig:protocol}
\end{figure*}

\begin{table}[t]
\caption{Hip exoskeleton hardware specifications.}
\label{tab:exo_hardware}
\centering \footnotesize \setlength{\tabcolsep}{4pt}
\begin{tabular}{p{0.42\columnwidth} p{0.48\columnwidth}}
\toprule
\textbf{Item} & \textbf{Specification} \\
\midrule
Assistance plane & Sagittal (hip flex/ext) \\
Active DOF & 1 per side (flexion/extension) \\
Passive DOF & 1 per side (abd/add) \\
Actuator & MyActuator X8-25 BLDC ($\times$2) \\
Peak torque & 25\,Nm per hip \\
Total mass & 5.94\,kg \\
Onboard computer & Raspberry Pi 5 \\
Joint sensing & Motor encoders or 3$\times$ IMU (LPMS-B2, LP-Research) \\
Command / logging rate & 50\,Hz / 50\,Hz \\
Communication & CAN bus (actuators), BLE (IMUs) \\
Materials & Aluminum, carbon tube, Onyx (Nylon-CF) \\
\bottomrule
\end{tabular}
\end{table}

\subsection{Hip Exoskeleton Hardware}
\label{subsec:hardware}
A bilateral hip exoskeleton was custom designed to provide sagittal-plane assistance (Fig.~\ref{fig:hardware}) with specifications listed in Table~\ref{tab:exo_hardware}. Each side has an actuated flexion/extension joint (MyActuator X8-25, 25\,Nm peak) and a passive abduction/adduction joint for fitting and lateral compliance. The waist frame adjusts in width to fit different pelvis sizes, and the thigh links adjust in length for different leg segments. The lateral extensions of the waist frame slide anteroposteriorly and the vest height adjusts vertically, so the actuator center can be aligned with each wearer's hip joint center (Fig.~\ref{fig:hardware}). A carbon-fiber back plate conforms to the wearer's back, which stabilizes the device against the torso and reduces wobbling during walking. The device's passive abduction/adduction degree of freedom is provided by a hinge between the waist frame and the actuator assembly, which lets the actuator swing with the thigh during lateral hip motion. Each participant was fitted before familiarization and settings were left unchanged for the remainder of the experiment. Bilateral hip kinematics can be obtained either from the built-in motor encoders or from three wireless IMUs (LPMS-B2, LP-Research, Tokyo, Japan) mounted on the pelvis and both thighs. The trained exoskeleton controller ran on an onboard Raspberry~Pi~5, which does not provide deterministic real-time scheduling. To improve timing reliability for near real-time control, both motors were commanded over the CAN bus at a moderate frequency of 50\,Hz (the same frequency used in simulation training). At a 20\,ms control period, the timing jitter introduced by the operating system remained negligible within each control cycle, while the lower update rate also reduced CAN bus utilization, contributing to more reliable communication.

\subsection{Sim-to-Real Deployment}
\label{subsec:sim2real}
The exoskeleton control policy (actor) trained in Stage~4 was deployed on hardware without subject-specific retraining or parameter re-tuning, and the on-device pipeline reproduced the simulated observation and action interfaces exactly. Hip angle was taken as the pelvis--femur difference from the three IMUs, and hip angular velocity as the difference between the thigh and pelvis gyroscope signals about the flexion axis. Before each trial, the IMUs were zeroed in a standing posture. At the 50\,Hz control rate, the actor consumed the same 18-dimensional observation as in training, comprising a three-step history of bilateral hip angles, angular velocities, and its own normalized torque outputs. It produced a normalized torque in $[-1,1]$ per hip, which was scaled by the preset torque limit and sent to the actuators over the CAN bus. The control loop was synchronized to a fixed 20\,ms period, with torque commands transmitted at the end of each cycle and the controller idle until the next update if computation completed early. Across all trials the controller maintained the intended 50\,Hz schedule without noted timing overruns. Because the policy was trained on unfiltered kinematics, no online filtering was applied at deployment, so that its input statistics on hardware matched those seen during training. A software torque limit and an emergency stop under continuous operator supervision were applied throughout all trials.

\subsection{Human Validation Framework}
\label{subsec:protocol}
The human study was designed to quantify the metabolic penalty of wearing the device, to quantify the benefit of its active assistance, and to assess whether a single policy generalizes across subjects without per-subject tuning. Eight healthy adults (5 male, 3 female; age $27.0\pm7.8$\,yr, range 18--44; mass $70.4\pm15.0$\,kg; height $1.73\pm0.06$\,m) participated under IRB approval (Protocol No.\ 2305033091R003) with written informed consent. The same subject numbering S1--S8 is used throughout, including the per-subject tables and the generalization analyses. Individual participant characteristics are listed in Table~S1 of the supplementary material.

Before data collection, each participant completed a familiarization session on the treadmill, walking 5\,min with the device unpowered and 10\,min with SMAT assistance active. Each subject then wore the hip exoskeleton and completed an order-balanced walking protocol that allows the metabolic penalty of the device and the benefit of its active assistance to be quantified separately (Fig.~\ref{fig:protocol}). Full order was standing rest, no-exoskeleton walking, passive walking, the SMAT and Ctrl-B (described below) controllers in randomized order, and then passive walking, no-exoskeleton walking, and standing rest. In the passive condition the device was worn with its commanded torque set to zero and its actuators back-drivable.

The two active trials comprised the SMAT controller and an alternative learning-based controller (\emph{Ctrl-B}) trained by a separate method with rewards based on muscle metabolic energy. Because the SMAT policy saturates its normalized output near $0.8$ rather than $1$ (shaped by the Stage~4 penalty, $\delta=0.8$), it was commanded at 15\,Nm so that its actual peak ($0.8\times15\approx12$\,Nm) matched the 12\,Nm commanded to Ctrl-B, placing both active trials at a common $\sim$12\,Nm assistance level. Although trained differently, the assistance Ctrl-B provided was similar in timing to that of the SMAT controller, and most users did not notice a drastic change between them. Additional details and results on Ctrl-B will be summarized in a future study, so its performance is not reported or compared here. Its inclusion in the protocol placed the two active trials in randomized order, which controls for order effects within the active block. 
All analyses in this article concern the No-Exo, Passive, and SMAT conditions.

The symmetric (order-balanced) condition ordering, in which the standing-rest, no-exo, and passive conditions were measured both before and after the active trials, controls for fatigue and temporal drift. The reported value for each of these conditions is the mean of its two repeats. Each walking trial lasted approximately 6\,min, with metabolic variables averaged over the final 2\,min to capture steady state. All walking was performed on a treadmill at 1.25\,m/s, and expired gases were recorded with a portable indirect calorimetry system (COSMED K5, COSMED, Italy). At least 5\,min of seated rest was provided between walking conditions, extended as needed until the participant felt ready to continue, to limit carry-over fatigue. All eight subjects completed this primary protocol. Two supplementary tests, across walking speeds and terrains, further assessed whether the single deployed policy generalizes beyond these training conditions.

\subsection{Metabolic Measurement}
\label{subsec:metabolic}
\looseness=-1
Whole-body metabolic rate was measured by indirect calorimetry~\cite{seo2016fully}. Instantaneous metabolic power was computed from oxygen uptake $\dot V_{O_2}$ and carbon-dioxide production $\dot V_{CO_2}$ using a Brockway-type relation~\cite{brockway1987derivation}:
\begin{equation}
\dot E = \frac{0.278\,\dot V_{O_2} + 0.075\,\dot V_{CO_2}}{m},
\label{eq:brockway}
\end{equation}
where $\dot V_{O_2}$ and $\dot V_{CO_2}$ are in mL/min, $m$ is body mass (kg), and $\dot E$ is the mass-specific metabolic rate (W/kg). $\dot E$ was averaged over the final 2\,min of each trial, following common practice for treadmill metabolic measurement. For the standing-rest, no-exo, and passive conditions, the two repeats were averaged. \emph{Net} metabolic rate was then obtained by subtracting the averaged standing-rest baseline, $\dot E_{\text{net}} = \dot E_{\text{cond}} - \dot E_{\text{rest}}$, and two physiologically meaningful contrasts were formed: the device \emph{penalty} (Passive vs.\ No-Exo) and the assistance \emph{benefit} (SMAT vs.\ Passive; SMAT vs.\ No-Exo).

\subsection{Biomechanical and Statistical Analysis}
\label{subsec:analysis}

Bilateral hip angle, angular velocity, and commanded torque were recorded during walking. 
Assistance mechanical power was computed as $P=\tau\,\omega$ after One-Euro filtering~\cite{casiez20121} of the angular velocity, and gait cycles were segmented at peak hip flexion, used as the reference event in the absence of ground-reaction-force sensors. To characterize the mechanical output of the controller, four per-subject metrics were computed following~\cite{lim2023parametric}: root-mean-square torque ($\tau_\text{RMS}$, Nm), mean peak torque ($\tau_\text{MAX}$, Nm), mean positive power (MPP, W), and mean negative power (MNP, W). For each gait cycle,
\begin{equation}
\text{MPP} = \frac{1}{n}\!\sum_{\tau_i\dot{q}_i>0}\!\tau_i\dot{q}_i,
\quad
\text{MNP} = \frac{1}{n}\!\sum_{\tau_i\dot{q}_i<0}\!\tau_i\dot{q}_i,
\label{eq:mpp_mnp}
\end{equation}
where $\tau_i$ and $\dot{q}_i$ are the hip torque and angular velocity and $n$ is the number of samples in the cycle. All metrics were then averaged across cycles within each subject.

Percentage changes in net metabolic rate were computed per subject and averaged across participants (mean of individual ratios). Group comparisons used two-tailed paired $t$-tests ($n=8$, $\text{df}=7$), applied after confirming that the paired differences did not deviate from normality under the Shapiro--Wilk test. For net metabolic rate, three comparisons were evaluated (Passive vs.\ No-Exo, SMAT vs.\ Passive, SMAT vs.\ No-Exo); for the hip kinematic and spatiotemporal variables (range of motion, peak flexion and extension, cadence, and stride length), six comparisons were evaluated. $p$-values were Bonferroni-corrected within each family and reported as corrected values, with significance at $\alpha=0.05$.

\section{Results}
\label{sec:results}

\subsection{Simulation: Stable Gait and Co-Adapted Assistance}
\label{subsec:sim_gait}

Stage~1 produced a stable human walking policy in approximately 320\,M simulation steps, with strongly symmetric joint kinematics (Pearson $r>0.92$) and a toe-off at $57.8\pm1.1\,\%$ of the gait cycle, consistent with treadmill-walking norms~\cite{perry2024gait}. Starting from this policy, Stage~2 accommodated the added device mass and inertia within roughly 160\,M steps while preserving the learned trajectories (RMSE $<3.6^\circ$ across all joints) and bilateral symmetry~\cite{forczek2012evaluation}.

\begin{figure}[t]
\centering
\includegraphics[width=1.0\columnwidth]{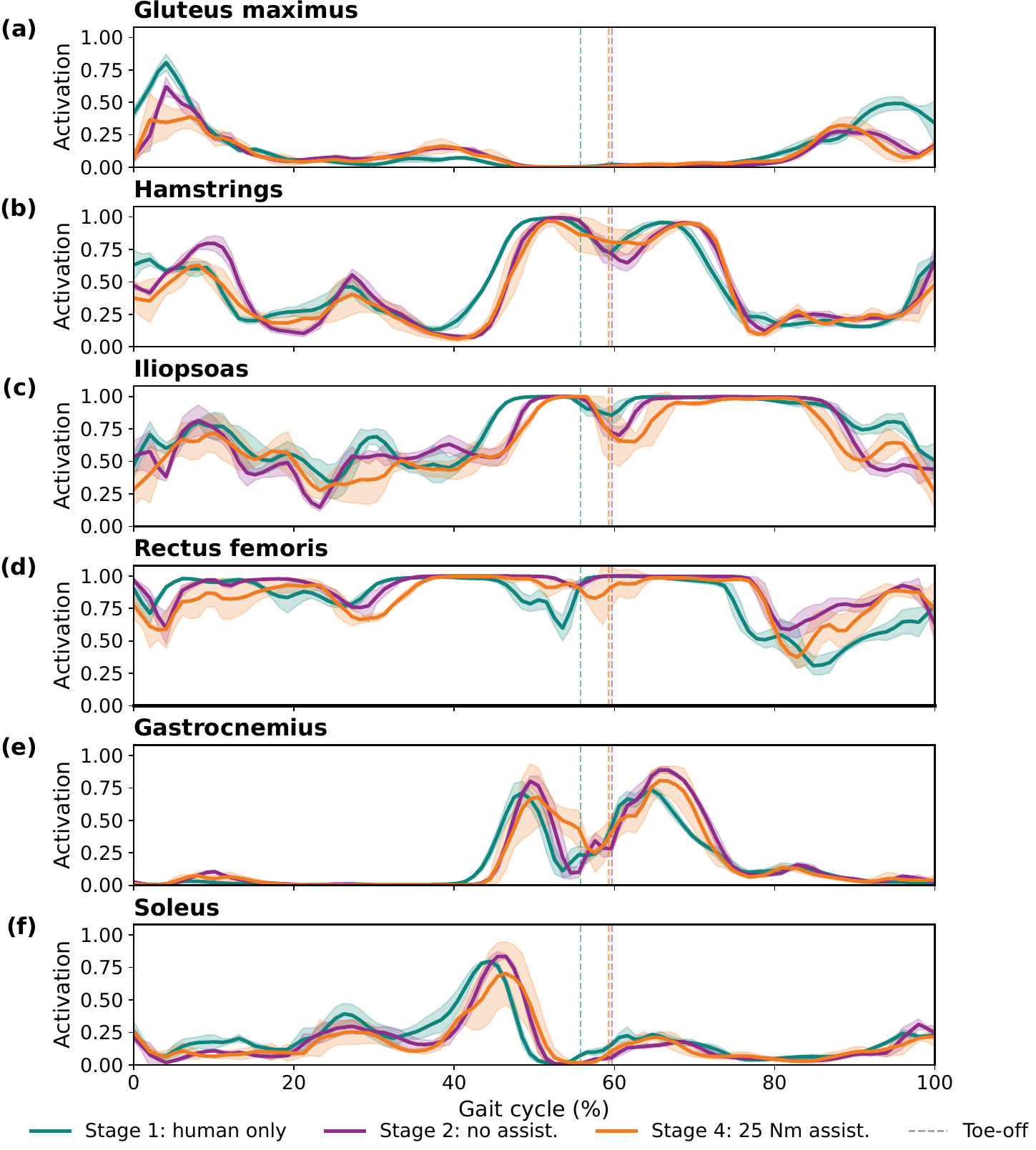}
\caption{Bilateral-averaged activation of six muscle groups (gluteus maximus, hamstrings, iliopsoas, rectus femoris, gastrocnemius, soleus) over the gait cycle for Stage~1 (human only, green), Stage~2 (exoskeleton attached, no assist, purple), and Stage~4 (25\,Nm assist, orange). Bands: mean $\pm$1\,s.d. Dashed lines: mean toe-off.}
\label{fig:muscle}
\end{figure}

Stage~3 established an initial flexion-assist timing pattern in about 1.1\,M steps, and Stage~4 refined it through full co-adaptation over approximately 66\,M steps. The resulting assistance torque peaks at a normalized value of $0.83$ during late-swing flexion ($75\,\%$ of the gait cycle) and confines negative work to only $10\,\%$ of the cycle in simulation (the portion where $P=\tau\dot{q}<0$), indicating that the delivered torque assists rather than opposes hip motion through nearly the entire stride. Whereas our earlier report~\cite{yuan2026smat} quantified one-side activation over the four hip muscle groups, here we average both legs across six muscle groups spanning the hip, knee, and ankle. Relative to Stage~2, in which the exoskeleton is worn without active assistance, Stage~4 assistance lowered activation across all six groups by $6.1\,\%$ on average (Fig.~\ref{fig:muscle}).

\subsection{Ablation: Necessity of Stages~3 and~4}
\label{subsec:ablation}

\begin{figure}[t]
\centering
\includegraphics[width=0.9\columnwidth]{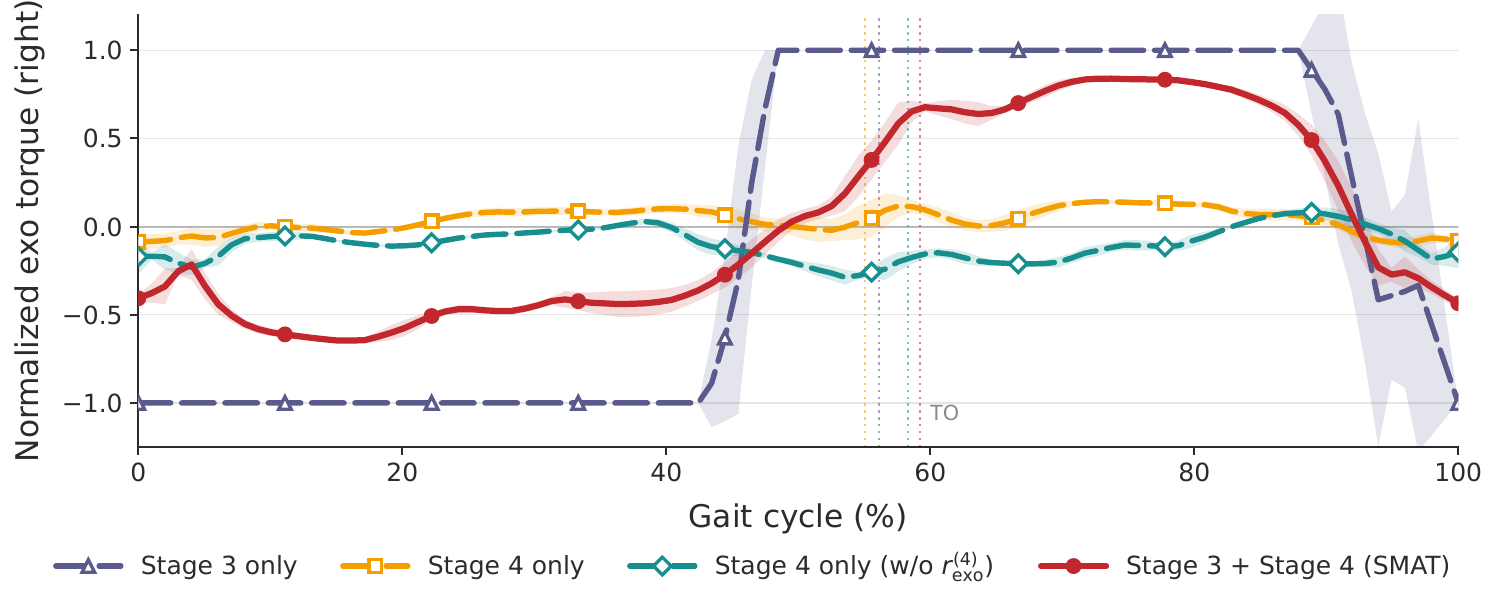}
\caption{Ablation of right-hip normalized exoskeleton torque over the gait cycle: \emph{Stage~3 only} (purple); \emph{Stage~4 only} from Stage~2 (orange); \emph{Stage~4 only without} $r_\mathrm{exo}^{(4)}$ (green); and the full \emph{Stage~3~+~Stage~4 (SMAT)} pipeline (red). Dotted vertical lines mark per-condition mean toe-off.}
\label{fig:ablation}
\end{figure}

As shown in Fig.~\ref{fig:ablation}, three ablation conditions reveal the necessity of both stages. Without Stage~4 co-adaptation (\emph{Stage~3 only}), the torque saturates at $\pm\hat{\tau}_{\max}$ for most of the gait cycle and reverses abruptly near toe-off, imposing impulsive hip loading that would pose safety concerns in physical human--robot interaction~\cite{moreno2009analysis}. Without Stage~3 pre-training (\emph{Stage~4 only}), training collapses to a local optimum: the second term of~\eqref{eq:rexoreward_stage4} rewards minimizing torque magnitude, so the policy finds that producing no torque maximizes that reward component, yielding near-zero output (peak normalized torque $0.14$ vs.\ $0.83$, an $83\,\%$ reduction), with this minimal torque opposing joint motion over $32\,\%$ of the cycle (vs.\ $10\,\%$ in the full SMAT pipeline). Removing $r_\mathrm{exo}^{(4)}$ entirely (\emph{Stage~4 only without} $r_\mathrm{exo}^{(4)}$) causes the exoskeleton to converge to a slight constant extension torque, which persistently opposes hip flexion during swing and results in negative exoskeleton power for $40\,\%$ of the gait cycle. These three conditions confirm that both Stage~3 pre-training and Stage~4 co-adaptation are required components of the full SMAT pipeline.

\subsection{Hip Biomechanics and Assistance Power Delivery}
\label{subsec:biomechanics}

\begin{figure*}[t]
\centering
\includegraphics[width=\textwidth]{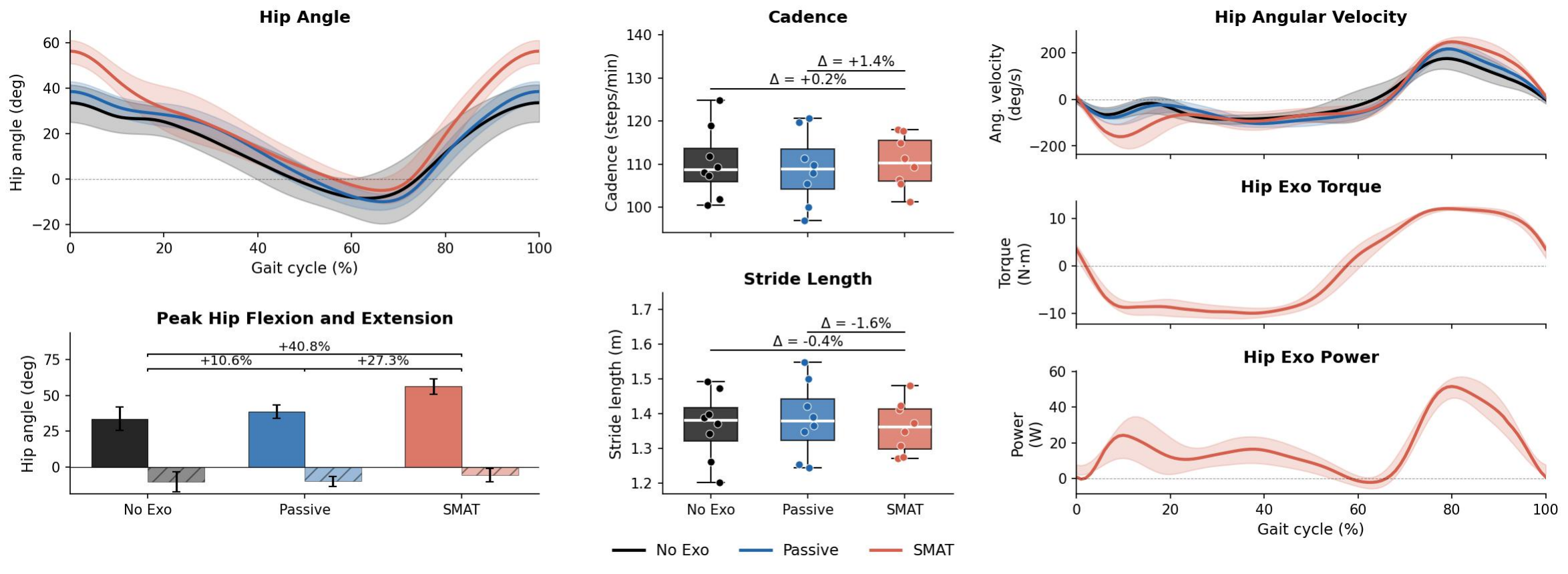}
\caption{Hip biomechanics across the No-Exo (black), Passive (blue), and SMAT (red) conditions during treadmill walking at 1.25\,m/s ($n=8$, both hips averaged per subject). \emph{Left:} hip angle over the gait cycle (flexion positive) and the corresponding peak hip flexion/extension, annotated with the percentage change in ROM. \emph{Middle:} cadence and stride length across conditions. \emph{Right:} hip angular velocity, SMAT exoskeleton torque, and exoskeleton mechanical power. Shaded bands and error bars: mean $\pm$1\,SD across subjects.}
\label{fig:biomech}
\end{figure*}

\looseness=-1
Fig.~\ref{fig:biomech} compares hip kinematics and exoskeleton delivered assistance power across the No-Exo, Passive, and SMAT conditions. The paired differences satisfied the Shapiro--Wilk normality check (all $p>0.05$). SMAT assistance markedly increased hip range of motion (ROM), from $44.0\pm4.3^\circ$ (No-Exo) to $62.0\pm5.5^\circ$ (SMAT), a $40.8\,\%$ increase (corrected $p<0.0001$), with the passive condition falling between the two ($48.7\pm4.0^\circ$, $+10.6\,\%$ vs.\ No-Exo, corrected $p=0.034$). This enlargement was driven primarily by flexion, which rose from $33.6\pm8.2^\circ$ to $56.3\pm5.2^\circ$ (corrected $p<0.001$), whereas peak extension did not change significantly ($-10.4\pm7.0^\circ$ to $-5.7\pm4.6^\circ$, corrected $p=0.47$). As absolute posture depends on the sensor zero, we quantify the change through ROM rather than the individual flexion and extension peaks. Since walking speed was fixed, cadence ($110.3\pm8.2$ vs.\ $110.5\pm6.1$\,steps/min, corrected $p>0.99$) and stride length ($1.366\pm0.098$ vs.\ $1.361\pm0.075$\,m, corrected $p>0.99$) were essentially unchanged between No-Exo and SMAT.
The corresponding peak angular velocities increased in the same order, reaching $250.9\pm18.9$\,deg/s in flexion and $-179.5\pm43.2$\,deg/s in extension under SMAT, while peak hip extension occurred at $63$--$67\,\%$ of the gait cycle in all three conditions, indicating that SMAT amplifies hip motion without shifting gait-event timing.
Across conditions, the group-mean hip-angle waveforms remained highly correlated in shape (Pearson $r=0.98$ for No-Exo vs.\ SMAT, $r=0.97$ for Passive vs.\ SMAT), confirming that the controller scales hip motion while preserving the normal gait pattern.

The exoskeleton delivered a bidirectional assistance torque, peaking at $12.1\pm0.2$\,N$\cdot$m of flexion assistance in late swing ($80\,\%$ of the gait cycle) with a slightly smaller extension-assistance phase of $10.3\pm0.8$\,N$\cdot$m near $25\,\%$. Mechanical power was predominantly positive and peaked near $52.8\pm4.0$\,W during flexion.
Table~\ref{tab:torque_power} reports the exoskeleton mechanical output in the SMAT condition for all eight subjects. RMS torque was $8.84\pm0.32$\,N$\cdot$m and mean peak torque $12.39\pm0.26$\,N$\cdot$m, consistent across subjects and matching the $\sim$12\,N$\cdot$m group-mean peak above. Mean positive power was $18.07\pm1.77$\,W, while mean negative power remained near zero ($-0.29\pm0.16$\,W) and the positive-power ratio was $R_\text{pos}=\text{MPP}/(\text{MPP}+|\text{MNP}|)=0.984 \pm 0.009$, confirming that the assistance was delivered almost entirely as positive power with minimal resistive losses. The low inter-subject variability in these metrics indicates that the single deployed policy produced consistent mechanical assistance across participants without subject-specific tuning.

\begin{table}[t]
\caption{Per-subject exoskeleton mechanical output, SMAT condition ($n=8$, both hips
averaged).}
\label{tab:torque_power}
\centering \setlength{\tabcolsep}{4pt}
\resizebox{\columnwidth}{!}{%
\begin{tabular}{lccccc}
\toprule
Subject & $\tau_\text{RMS}$ (N$\cdot$m) & $\tau_\text{MAX}$ (N$\cdot$m) & MPP (W) & MNP (W) & $R_\text{pos}$ \\
\midrule
S1 & 9.38 & 11.91 & 20.13 & $-0.37$ & 0.982 \\
S2 & 9.05 & 12.60 & 20.75 & $-0.30$ & 0.986 \\
S3 & 8.74 & 12.62 & 15.93 & $-0.35$ & 0.979 \\
S4 & 8.64 & 12.61 & 18.96 & $-0.29$ & 0.985 \\
S5 & 8.35 & 12.43 & 16.81 & $-0.60$ & 0.966 \\
S6 & 9.08 & 12.28 & 17.31 & $-0.07$ & 0.996 \\
S7 & 8.69 & 12.52 & 16.37 & $-0.21$ & 0.987 \\
S8 & 8.79 & 12.15 & 18.32 & $-0.17$ & 0.991 \\
\midrule
Mean$\pm$SD & $8.84 \pm 0.32$ & $12.39 \pm 0.26$ & $18.07 \pm 1.77$ & $-0.29 \pm 0.16$ & $0.984 \pm 0.009$ \\
\bottomrule
\end{tabular}}
\vspace{0.5mm}
\begin{minipage}{\columnwidth}\footnotesize
$\tau_\text{RMS}$: RMS torque; $\tau_\text{MAX}$: mean per-cycle peak torque; MPP/MNP: mean positive/negative power (Eq.~\eqref{eq:mpp_mnp}); $R_\text{pos}=\text{MPP}/(\text{MPP}+|\text{MNP}|)$: positive-power ratio. All metrics were computed
per gait cycle and averaged across cycles within each subject, with the two hips
averaged.
\end{minipage}
\end{table}

\subsection{Metabolic Reduction for Walking}
\label{subsec:metabolic_results}

\begin{figure}[t]
\centering
\includegraphics[width=0.92\columnwidth]{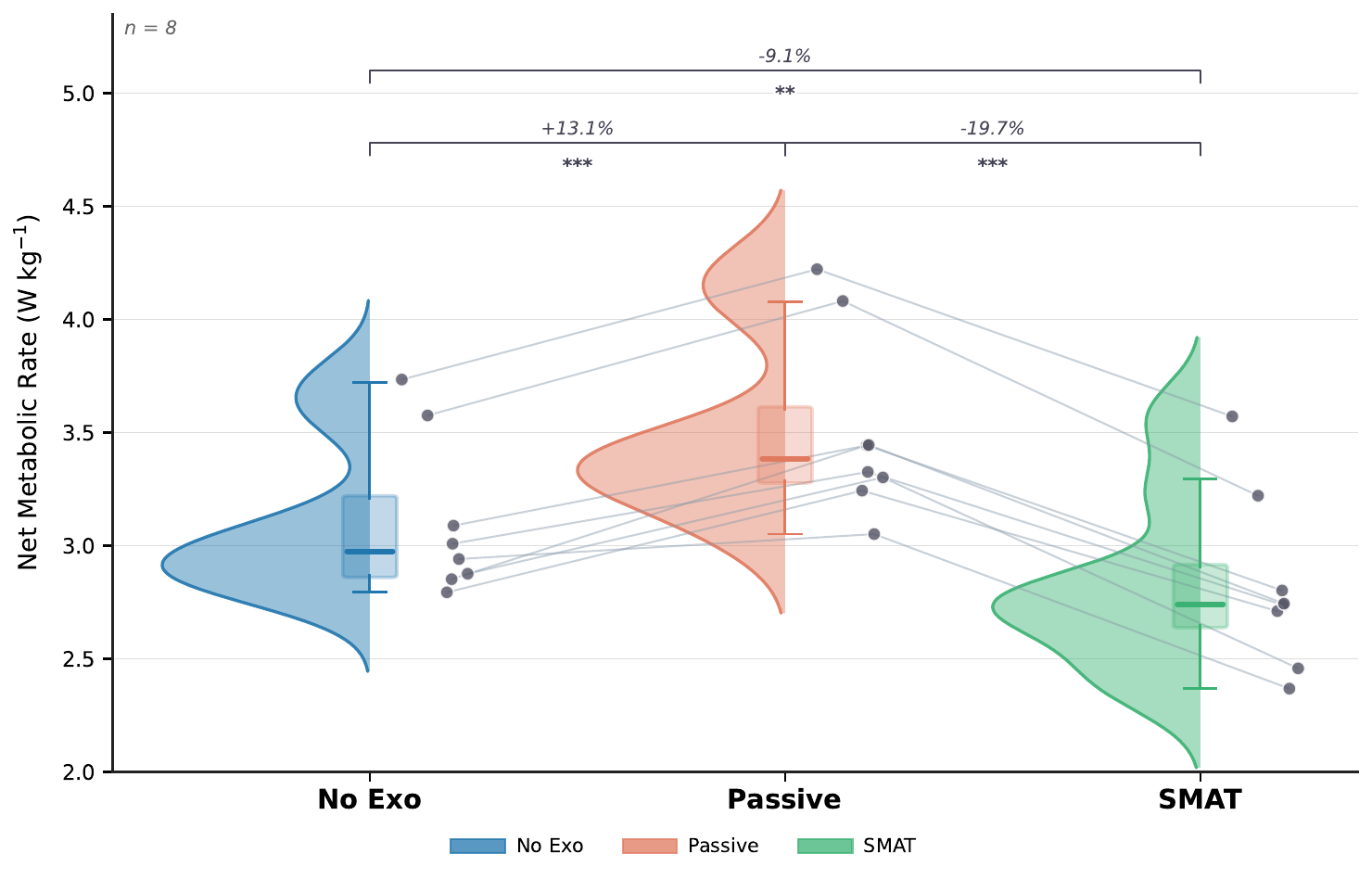}
\caption{Net metabolic rate ($\mathrm{W\,kg^{-1}}$) across conditions ($n=8$). Half-violin distributions, box plots, and individual subjects (connected lines) for No-Exo, Passive, and SMAT conditions. The half-violins show the full shape of each distribution, with density concentrated at lower metabolic rates under SMAT. Brackets show the group-mean percentage change between conditions ($+13.1\,\%$, $-19.7\,\%$, $-9.1\,\%$). Paired $t$-test with Bonferroni correction: $^{*}p<0.05$, $^{**}p<0.01$, $^{***}p<0.001$.}
\label{fig:metabolic}
\end{figure}

Net metabolic rate across the three conditions is summarized in Fig.~\ref{fig:metabolic} and Table~\ref{tab:metabolic}. The paired differences passed the Shapiro--Wilk normality check (all $p>0.4$). Wearing the exoskeleton passively raised net metabolic rate by $13.1\pm4.9\,\%$ relative to walking without the device ($t=7.80$, corrected $p<0.001$), confirming that the unpowered device imposes a metabolic penalty through its added mass and motion constraints. Active SMAT assistance lowered net metabolic rate by $19.7\pm3.3\,\%$ relative to the passive condition ($t=-17.00$, corrected $p<0.001$), more than offsetting this penalty. Relative to the no-exo condition, SMAT assistance reduced net metabolic rate by $9.1\pm6.0\,\%$ ($t=-4.50$, corrected $p<0.01$), with all eight participants lowering their cost under assistance (Fig.~\ref{fig:metabolic}, connected lines). The controller therefore did not merely compensate for the added burden of the hardware but produced a net reduction relative to unassisted walking, achieved with a single policy deployed across all subjects without subject-specific tuning.

Across all walking trials the respiratory exchange ratio (RER, $\dot V_{CO_2}/\dot V_{O_2}$) stayed below 1.0 ($0.84\pm0.03$, range 0.79--0.92; $0.83$ No-Exo, $0.84$ Passive, $0.85$ SMAT), confirming that all conditions were sustained aerobically and that Eq.~\eqref{eq:brockway} applies. The two repeats of the passive and no-exoskeleton conditions differed by $3.5\pm2.7\,\%$ and $6.9\pm5.2\,\%$, respectively, with no consistent direction across subjects. Both repeats were averaged for analysis, which removes any linear drift over the session.

\begin{table}[t]
\caption{Net metabolic rate (W/kg) per subject and condition ($n=8$), with the reduction under SMAT relative to the passive condition. $\Delta$ = Passive $-$ SMAT. Percentages are computed per subject and then averaged.}
\label{tab:metabolic}
\centering \setlength{\tabcolsep}{4pt}
\resizebox{\columnwidth}{!}{%
\begin{tabular}{lccccc}
\toprule
Subject & No-Exo & Passive & SMAT & $\Delta$ (Passive$-$SMAT) & Reduction (\%) \\
\midrule
S1 & 3.008 & 3.326 & 2.738 & 0.588 & 17.7 \\
S2 & 3.575 & 4.081 & 3.221 & 0.860 & 21.1 \\
S3 & 2.941 & 3.051 & 2.368 & 0.683 & 22.4 \\
S4 & 2.794 & 3.243 & 2.711 & 0.532 & 16.4 \\
S5 & 3.734 & 4.222 & 3.571 & 0.651 & 15.4 \\
S6 & 2.876 & 3.301 & 2.458 & 0.843 & 25.5 \\
S7 & 2.852 & 3.444 & 2.802 & 0.642 & 18.6 \\
S8 & 3.088 & 3.445 & 2.744 & 0.701 & 20.3 \\
\midrule
Mean$\pm$SD & $3.11\pm0.35$ & $3.51\pm0.41$ & $2.83\pm0.39$ & $0.688\pm0.114$ & $19.7\pm3.3$ \\
\bottomrule
\end{tabular}}
\end{table}

\begin{figure}[t]
\centering
\includegraphics[width=\columnwidth]{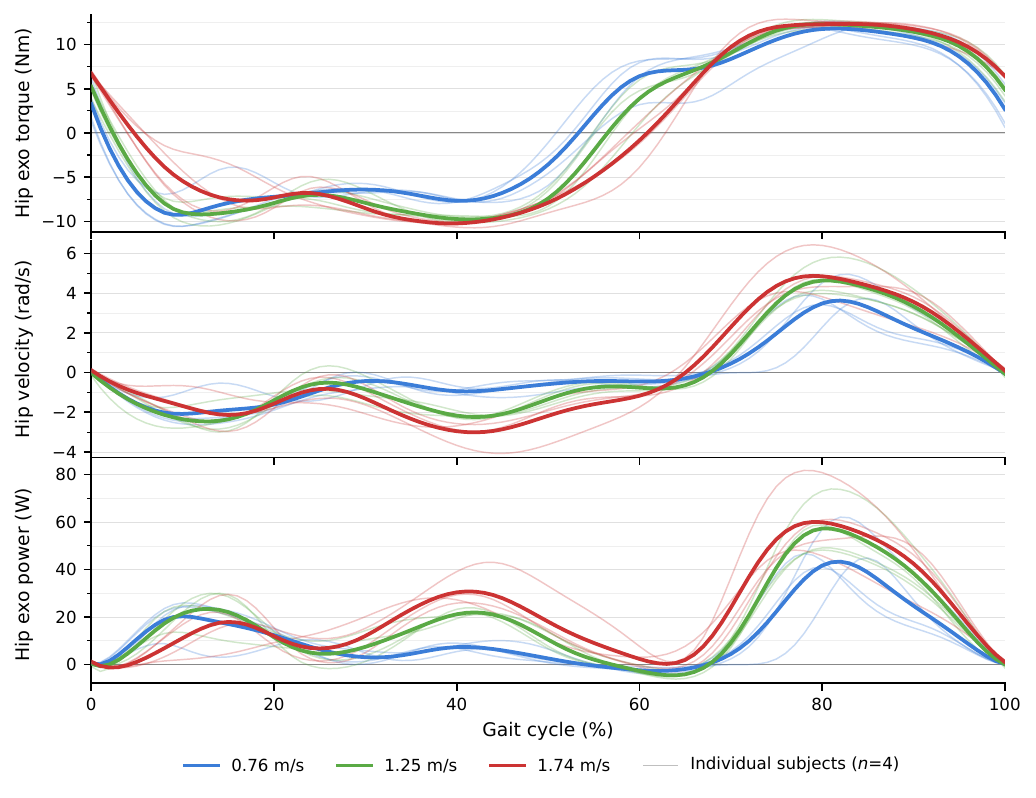}
\caption{Speed generalization of the SMAT controller at 0.76, 1.25, and 1.74\,m/s (blue/green/red; $n=4$, trained at 1.25\,m/s only). \emph{Top:} gait-cycle-normalized exo torque. \emph{Middle:} hip angular velocity. \emph{Bottom:} exo mechanical power. Thin lines: individual-subject means; bold: group mean. Gait cycles segmented at peak hip flexion.}
\label{fig:speed_gen}
\end{figure}

\begin{figure*}[t]
\centering
\includegraphics[width=\textwidth]{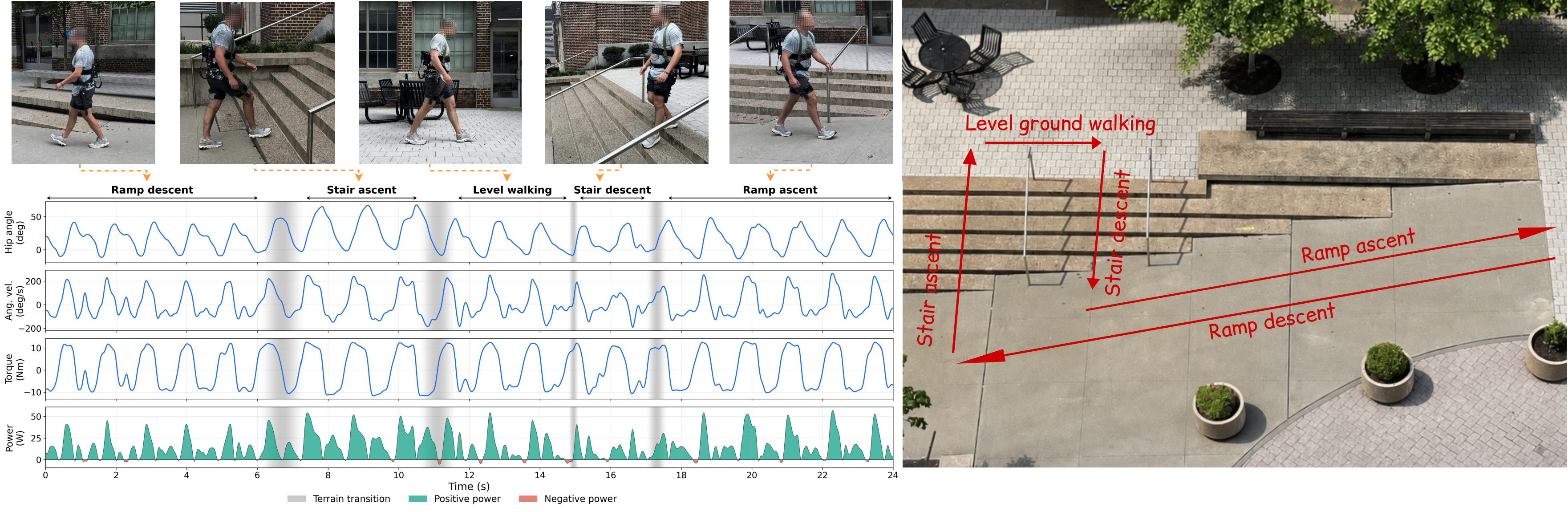}
\caption{Terrain generalization of the SMAT controller during a continuous outdoor circuit (participant S3). \emph{Left:} hip angle, angular velocity, commanded torque, and exoskeleton mechanical power across five modes (ramp descent, stair ascent, level walking, stair descent, and ramp ascent), with terrain transitions shaded. \emph{Right:} the outdoor test course. The same Stage~4 policy was used throughout, with no terrain-specific switching.}
\label{fig:terrain}
\end{figure*}

\begin{table*}[t]
\caption{Per-Subject Exoskeleton Mechanical Power Across Walking Speeds}
\label{tab:speed_gen}
\centering
\renewcommand{\arraystretch}{1.05}
\setlength{\tabcolsep}{4pt}
\resizebox{\textwidth}{!}{%
\begin{tabular}{l|ccc|ccc|ccc}
\hline
 & \multicolumn{3}{c|}{0.76\,m/s}
 & \multicolumn{3}{c|}{1.25\,m/s$^\dagger$}
 & \multicolumn{3}{c}{1.74\,m/s} \\
\cline{2-4}\cline{5-7}\cline{8-10}
Subject & MPP (W) & MNP (W) & $R_\text{pos}$ & MPP (W) & MNP (W) & $R_\text{pos}$ & MPP (W) & MNP (W) & $R_\text{pos}$ \\
\hline
S1 & 11.05 & $-$0.14 & 0.988 & 15.95 & $-$0.49 & 0.970 & 20.20 & $-$0.17 & 0.992 \\
S2 & 15.53 & $-$0.42 & 0.974 & 22.16 & $-$0.46 & 0.980 & 28.35 & $-$0.08 & 0.997 \\
S3 &  8.14 & $-$0.30 & 0.964 & 15.80 & $-$0.19 & 0.988 & 17.26 & $-$0.09 & 0.995 \\
S5 & 11.83 & $-$0.37 & 0.970 & 17.40 & $-$0.57 & 0.968 & 20.04 & $-$0.31 & 0.985 \\
\hline
Mean ($\pm$SD)
   & $11.6\pm3.0$ & $-0.31\pm0.12$ & $0.974\pm0.010$
   & $17.8\pm3.0$ & $-0.43\pm0.16$ & $0.977\pm0.009$
   & $21.5\pm4.8$ & $-0.16\pm0.11$ & $0.992\pm0.005$ \\
\hline
\end{tabular}}
\vspace{0.5mm}
\begin{minipage}{\textwidth}\footnotesize
MPP/MNP: mean positive/negative power (Eq.~\eqref{eq:mpp_mnp}); $R_\text{pos} = \text{MPP}/(\text{MPP} + |\text{MNP}|)$: positive-power ratio. All trials: same torque limit across speeds, peak assistance $\approx$12\,Nm. $^\dagger$Training speed. Right leg, cycles segmented at peak hip flexion.
\end{minipage}
\end{table*}

\subsection{Generalization Across Walking Speeds}
\label{subsec:speed_gen}
We further tested how the SMAT controller responds to walking speeds beyond its trained velocity at 1.25\,m/s. In a separate session, four participants (S1, S2, S3, S5) walked at 0.76, 1.25, and 1.74\,m/s on a treadmill under SMAT assistance, using the same exoskeleton and controller.
Fig.~\ref{fig:speed_gen} shows the gait-cycle-normalized exo torque, hip angular velocity, and mechanical power across speeds. With the torque limit held constant across speeds (peak assistance $\approx$12\,Nm), mean positive power (MPP) scaled with walking speed, reaching $11.6\pm3.0$\,W at 0.76\,m/s, $17.8\pm3.0$\,W at 1.25\,m/s, and $21.5\pm4.8$\,W at 1.74\,m/s, driven by the increase in hip angular velocity at faster speed and a slight timing shift of the torque profile. Mean negative power remained below 0.6\,W in magnitude and the positive-power ratio ($R_\text{pos}$) exceeded 0.96 at all speeds, indicating that the torque was predominantly assistive throughout the stride. Per-subject results are listed in Table~\ref{tab:speed_gen}. The torque peak consistently occurred during late-swing hip flexion across all three speeds without any speed-based switching, suggesting the policy learned a kinematics-driven assistive pattern rather than a cadence-locked one. The same controller thus transfers across walking speeds from $0.76$ to $1.74$\,m/s without per-subject retraining or explicit speed adaptation.

\subsection{Generalization Across Terrains}
\label{subsec:terrain}
Although the controller was trained on level ground only, we assessed its stability on terrains well outside this training distribution. As a feasibility demonstration, one participant (S3) completed a continuous outdoor circuit spanning ramp descent, stair ascent, level walking, stair descent, and ramp ascent, without stopping between modes and without any terrain-specific switching or gait-phase classification. The same policy and near 12\,N$\cdot$m peak assistance from the treadmill study were used throughout. Fig.~\ref{fig:terrain} shows the hip angle, angular velocity, commanded torque, and mechanical power over the circuit. The assistance remained smooth through every transition, with no torque overshoot at mode boundaries and power that stayed almost entirely positive across all modes ($R_\text{pos} = 0.96$ to $1.00$). No destabilizing behavior or safety intervention occurred. Because the policy receives only hip kinematics and assumes no gait phase or terrain, it adapted to each mode through the same kinematics-driven behavior identified in Section~\ref{subsec:speed_gen}. 

\section{Discussion}
\label{sec:discussion}
The central finding of this study is the successful demonstration of the SMAT framework for obtaining a unified hip-exoskeleton assistance policy and validating its physiological benefit. Active SMAT assistance significantly reduced net metabolic rate relative to wearing the device passively ($-19.7\,\%$), more than compensating for the $+13.1\,\%$ metabolic penalty imposed by the passive device, and reduced net cost below the no-exo condition by $9.1\,\%$ ($p<0.01$). Notably, this benefit was obtained with a single policy deployed across all subjects without subject-specific retraining, supporting the practical value of the staged co-adaptive training approach. 
Comparing the active and passive exoskeleton conditions isolates the pure assistance benefit from the metabolic penalty associated with wearing the device, whereas the reduction relative to the no-exo condition reflects the net benefit of assistance after accounting for the additional hardware burden. These results further suggest that using a device with a lower inherent penalty could enable an even greater reduction in metabolic cost relative to the no-exo condition.

Several features of the assistance help explain this reduction. The controller produced a bidirectional torque that assisted both the flexion and extension phases rather than a single burst, and its mechanical power was almost entirely positive, with a positive-power ratio of $0.984 \pm 0.009$. Each watt of positive mechanical power corresponded to a net metabolic saving of about $1.3$\,W relative to the passive condition (average $47.9\pm10.4$\,W of metabolic saving against $36.1\pm3.5$\,W of total positive power from both legs). 

The metabolic reduction achieved in this study compares favorably with or falls within the range of values reported for recent rigid hip exoskeletons \cite{sawicki2020exoskeleton}, while offering the practical advantage of a controller trained entirely in simulation and deployed across subjects without per-subject calibration. For example, under a similar protocol at $1.25$\,m/s, a data-driven, neural-network-controlled hip exoskeleton that autonomously estimated joint moments lowered metabolic cost by $12.8\,\%$ relative to a zero-torque condition without user-specific tuning~\cite{molinaro2024task}. 
Two recent studies have also demonstrated the potential of simulation-trained hip controllers for reducing metabolic cost. Luo \emph{et al.}~\cite{luo2024experiment} and Barati \emph{et al.}~\cite{barati2026end} report $24.3\,\%$ and $15.2\,\%$ reductions at $1.25$\,m/s, respectively, against no-exo conditions.  These studies used much lighter and less encumbering devices weighing $3.2$ and $2.9$\,kg, respectively, compared with the $5.94$\,kg device used in the present study, with respective peak torques of $18$\,N$\cdot$m and $11$\,N$\cdot$m. An autonomous hip exoskeleton with an even lighter $2.8$\,kg device likewise reduced cost by $13\,\%$ against not wearing the device~\cite{seo2016fully}. The reference condition and device burden are important when interpreting these comparisons. Relative to the no-exo condition, the $9.1\,\%$ reduction observed in this study absorbs the penalty of our heavier hardware: the unpowered mass raised net cost by $13.1\,\%$, against only $+5.4\,\%$ for~\cite{luo2024experiment} (Barati \emph{et al.}~\cite{barati2026end} did not report the weight penalty). This penalty is consistent with reports that a heavier hip exoskeleton can raise metabolic cost well above the no-exo condition~\cite{kang2019effect}. The active assistance from our SMAT controller more than offsets this substantial metabolic penalty, while the active-versus-passive comparison isolates the benefit of assistance from the metabolic burden of the device. Together with the absence of subject-specific retraining, these results demonstrate the potential of simulation-trained co-adaptive control to provide substantial physiological benefits across users.

Wearing the unpowered device raised whole-body metabolic cost, but in simulation several hip muscles were less active at Stage~2 than at Stage~1. Examining the knee and ankle muscles across the two stages, activation of the gastrocnemius and several other distal muscles rose while that of soleus fell, indicating that the added mass reorganized how the muscles were recruited. Whether real users show the same change, and whether it matches simulation, remains to be verified with future EMG studies.

Our controller ran at only 50\,Hz on a Raspberry~Pi, reaching a high level of effectiveness on modest embedded hardware without high-rate control. Deploying the policy on IMU-derived hip kinematics rather than motor-encoder measurements helped preserve the assistance timing learned in simulation, since inertial sensing tracks limb motion more directly and is less affected by soft-tissue compliance and strap looseness. 
We did observe mild inter-limb asymmetry, likely from the absence of an explicit symmetry constraint during training, which a bilateral symmetry term could regularize in future work.

\looseness=-1
There are several limitations in the current study. The study involved eight healthy adults walking on a treadmill at a single speed, so the metabolic benefit remains to be confirmed in larger cohorts, in overground walking, and across a broader range of speeds. We also recorded only hip kinematics. Hip range of motion increased under assistance, which provides evidence of altered human--exoskeleton coordination consistent with co-adaptation. However, cadence, stride length, and extension timing were mostly unchanged, so the enlarged motion occurred within the same gait structure, but how it is accommodated at the knee and ankle was not quantified. Capturing full lower-limb kinematics would clarify this. Our training was conducted within a single musculoskeletal simulation framework (MyoAssist open source framework \cite{tan2025myoassist}), in which both the human and the exoskeleton policies are learned to co-adapt. Although the staged curriculum effectively managed the non-stationarity arising from mutual adaptation in this setting, the extent to which the specific four-stage decomposition generalizes to other simulation frameworks and learning configurations remains to be established. 
Future studies comparing SMAT with alternative curriculum designs and training approaches across different musculoskeletal simulation frameworks would help establish the broader applicability of multi-stage co-adaptive training. Beyond the level-walking task considered here, extending SMAT to more demanding activities such as stair climbing, sit-to-stand transitions, and sloped-terrain walking could provide further opportunities to demonstrate the value of multi-stage training and human--exoskeleton co-adaptation. These tasks introduce more complex changes in locomotor dynamics and assistance requirements, potentially making the benefits of structured, staged adaptation even more pronounced. Such extensions represent a promising direction toward developing versatile co-adaptive controllers for a broader range of real-world locomotor conditions.

\section{Conclusion}
\label{sec:conclusion}
We demonstrated that SMAT can produce a simulation-trained human--exoskeleton co-adaptive policy that transfers to real users without subject-specific retraining. A single final policy deployed on a hip exoskeleton reduced net metabolic rate by 19.7\,\% relative to the passive device ($p<0.001$) across eight healthy adults. Joint-level biomechanical analysis further showed that the policy delivered predominantly positive and appropriately timed hip mechanical power (positive-power ratio $0.984$) and markedly increased hip range of motion, providing evidence of altered human--exoskeleton coordination consistent with co-adaptation. The same policy maintained predominantly positive assistance across walking speeds from 0.76 to 1.74\,m/s and across outdoor terrains beyond the training conditions. Together with the simulation results and staged ablation experiments, these findings demonstrate the potential of simulation-based co-adaptive learning to produce transferable exoskeleton assistance while achieving a significant reduction in metabolic cost and reducing the experimental burden of subject-specific controller optimization.

\section*{Acknowledgment}
This work was partially supported by the National Institute on Disability, Independent Living, and Rehabilitation Research (NIDILRR) funded Rehabilitation Engineering Research Center Grant 90REGE0025-01-00 and NSF Award \#2524089.

\balance
\bibliographystyle{IEEEtran}
\bibliography{reference_arxiv}

@article{luo2023robust,
  title={Robust walking control of a lower limb rehabilitation exoskeleton coupled with a musculoskeletal model via deep reinforcement learning},
  author={Luo, Shuzhen and Androwis, Ghaith and Adamovich, Sergei and Nunez, Erick and Su, Hao and Zhou, Xianlian},
  journal={Journal of neuroengineering and rehabilitation},
  volume={20},
  number={1},
  pages={34},
  year={2023},
  publisher={Springer}
}

@article{luo2024experiment,
  title={Experiment-free exoskeleton assistance via learning in simulation},
  author={Luo, Shuzhen and Jiang, Menghan and Zhang, Sainan and Zhu, Junxi and Yu, Shuangyue and Dominguez Silva, Israel and Wang, Tian and Rouse, Elliott and Zhou, Bolei and Yuk, Hyunwoo and others},
  journal={Nature},
  volume={630},
  number={8016},
  pages={353--359},
  year={2024},
  publisher={Nature Publishing Group UK London}
}

@inproceedings{tan2025myoassist,
  title = {{MyoAssist} 0.1: {MyoSuite} for Dexterity and Agility in Bionic Humans},
  author={Tan, Chun Kwang and Wang, Cheryl and Lyu, Shirui and Hodossy, Balint K and Schumacher, Pierre and Wilson, Elizabeth B and Caggiano, Vittorio and Kumar, Vikash and Farina, Dario and Gionfrida, Letizia and others},
  booktitle={2025 International Conference On Rehabilitation Robotics (ICORR)},
  pages={437--442},
  year={2025},
  organization={IEEE}
}

@article{slade2024human,
  title={On human-in-the-loop optimization of human--robot interaction},
  author={Slade, Patrick and Atkeson, Christopher and Donelan, J Maxwell and Houdijk, Han and Ingraham, Kimberly A and Kim, Myunghee and Kong, Kyoungchul and Poggensee, Katherine L and Riener, Robert and Steinert, Martin and others},
  journal={Nature},
  volume={633},
  number={8031},
  pages={779--788},
  year={2024},
  publisher={Nature Publishing Group UK London}
}

@article{chiappa2024acquiring,
  title={Acquiring musculoskeletal skills with curriculum-based reinforcement learning},
  author={Chiappa, Alberto Silvio and Tano, Pablo and Patel, Nisheet and Ingster, Abiga{\"\i}l and Pouget, Alexandre and Mathis, Alexander},
  journal={Neuron},
  volume={112},
  number={23},
  pages={3969--3983},
  year={2024},
  publisher={Elsevier}
}

@article{molinaro2024task,
  title={Task-agnostic exoskeleton control via biological joint moment estimation},
  author={Molinaro, Dean D and Scherpereel, Keaton L and Schonhaut, Ethan B and Evangelopoulos, Georgios and Shepherd, Max K and Young, Aaron J},
  journal={Nature},
  volume={635},
  number={8038},
  pages={337--344},
  year={2024},
  publisher={Nature Publishing Group UK London}
}

@article{lim2023parametric,
  title={Parametric delayed output feedback control for versatile human-exoskeleton interactions during walking and running},
  author={Lim, Bokman and Choi, Byungjune and Roh, Changhyun and Hyung, Seungyong and Kim, Yong-Jae and Lee, Younbaek},
  journal={IEEE Robotics and Automation Letters},
  volume={8},
  number={8},
  pages={4497--4504},
  year={2023},
  publisher={IEEE}
}

@article{leem2026exo,
  title={{Exo-Plore}: Exploring Exoskeleton Control Space through Human-aligned Simulation},
  author={Leem, Geonho and Lee, Jaedong and Lee, Jehee and Song, Seungmoon and Won, Jungdam},
  journal={arXiv preprint arXiv:2601.22550},
  year={2026}
}

@article{yuan2026gait,
  title={Gait Asymmetry from Unilateral Weakness and Improvement With Ankle Assistance: a Reinforcement Learning based Simulation Study},
  author={Yuan, Yifei and Androwis, Ghaith and Zhou, Xianlian},
  journal={IEEE RAS/EMBS 11th International Conference on Biomedical Robotics and Biomechatronics (BioRob 2026), arXiv preprint arXiv:2602.18862},
  year={2026}
}

@article{ratnakumar2026reinforcement,
  title={Reinforcement-Learning-Based Assistance Reduces Squat Effort with a Modular Hip--Knee Exoskeleton},
  author={Ratnakumar, Neethan and Tohfafarosh, Mariya Huzaifa and Jauhri, Saanya and Zhou, Xianlian},
  journal={IEEE RAS/EMBS 11th International Conference on Biomedical Robotics and Biomechatronics (BioRob 2026), arXiv preprint arXiv:2602.17794},
  year={2026}
}

@article{sawicki2020exoskeleton,
  title={The exoskeleton expansion: improving walking and running economy},
  author={Sawicki, Gregory S and Beck, Owen N and Kang, Inseung and Young, Aaron J},
  journal={Journal of neuroengineering and rehabilitation},
  volume={17},
  number={1},
  pages={25},
  year={2020},
  publisher={Springer}
}

@article{rodriguez2021systematic,
  title={Systematic review on wearable lower-limb exoskeletons for gait training in neuromuscular impairments},
  author={Rodr{\'\i}guez-Fern{\'a}ndez, Antonio and Lobo-Prat, Joan and Font-Llagunes, Josep M},
  journal={Journal of neuroengineering and rehabilitation},
  volume={18},
  number={1},
  pages={22},
  year={2021},
  publisher={Springer}
}

@article{poggensee2021adaptation,
  title={How adaptation, training, and customization contribute to benefits from exoskeleton assistance},
  author={Poggensee, Katherine L and Collins, Steven H},
  journal={Science Robotics},
  volume={6},
  number={58},
  pages={eabf1078},
  year={2021},
  publisher={American Association for the Advancement of Science}
}

@inproceedings{casiez20121,
  title={1€ filter: a simple speed-based low-pass filter for noisy input in interactive systems},
  author={Casiez, G{\'e}ry and Roussel, Nicolas and Vogel, Daniel},
  booktitle={Proceedings of the SIGCHI Conference on human factors in computing systems},
  pages={2527--2530},
  year={2012}
}

@book{perry2024gait,
  title={Gait analysis: normal and pathological function},
  author={Perry, Jacquelin and Burnfield, Judith},
  year={2024},
  publisher={CRC Press}
}

@article{lowe2017multi,
  title={Multi-agent actor-critic for mixed cooperative-competitive environments},
  author={Lowe, Ryan and Wu, Yi I and Tamar, Aviv and Harb, Jean and Pieter Abbeel, OpenAI and Mordatch, Igor},
  journal={Advances in neural information processing systems},
  volume={30},
  year={2017}
}

@inproceedings{bengio2009curriculum,
  title={Curriculum learning},
  author={Bengio, Yoshua and Louradour, J{\'e}r{\^o}me and Collobert, Ronan and Weston, Jason},
  booktitle={Proceedings of the 26th annual international conference on machine learning},
  pages={41--48},
  year={2009}
}

@article{baud2021review,
  title={Review of control strategies for lower-limb exoskeletons to assist gait},
  author={Baud, Romain and Manzoori, Ali Reza and Ijspeert, Auke and Bouri, Mohamed},
  journal={Journal of neuroengineering and rehabilitation},
  volume={18},
  number={1},
  pages={119},
  year={2021},
  publisher={Springer}
}

@article{simos2025reinforcement,
  title={Reinforcement learning-based motion imitation for physiologically plausible musculoskeletal motor control},
  author={Simos, Merkourios and Chiappa, Alberto Silvio and Mathis, Alexander},
  journal={arXiv preprint arXiv:2503.14637},
  year={2025}
}

@article{forczek2012evaluation,
  title={An evaluation of symmetry in the lower limb joints during the able-bodied gait of women and men},
  author={Forczek, Wanda and Staszkiewicz, Robert},
  journal={Journal of human kinetics},
  volume={35},
  pages={47},
  year={2012}
}

@inproceedings{seo2016fully,
  title={Fully autonomous hip exoskeleton saves metabolic cost of walking},
  author={Seo, Keehong and Lee, Jusuk and Lee, Younbaek and Ha, Taesin and Shim, Youngbo},
  booktitle={2016 IEEE International Conference on Robotics and Automation (ICRA)},
  pages={4628--4635},
  year={2016},
  organization={IEEE}
}

@article{moreno2009analysis,
  title={Analysis of the Human Interaction with a Wearable Lower-Limb Exoskeleton},
  author={Moreno, Juan C and Brunetti, Fernando and Navarro, Enrique and Forner-Cordero, Arturo and Pons, Jos{\'e} L},
  journal={Applied Bionics and Biomechanics},
  volume={6},
  number={2},
  pages={245--256},
  year={2009},
  publisher={Wiley Online Library}
}

@article{zhang2017human,
  title={Human-in-the-loop optimization of exoskeleton assistance during walking},
  author={Zhang, Juanjuan and Fiers, Pieter and Witte, Kirby A and Jackson, Rachel W and Poggensee, Katherine L and Atkeson, Christopher G and Collins, Steven H},
  journal={Science},
  volume={356},
  number={6344},
  pages={1280--1284},
  year={2017},
  publisher={American Association for the Advancement of Science}
}

@article{ding2018human,
  title={Human-in-the-loop optimization of hip assistance with a soft exosuit during walking},
  author={Ding, Ye and Kim, Myunghee and Kuindersma, Scott and Walsh, Conor J},
  journal={Science robotics},
  volume={3},
  number={15},
  pages={eaar5438},
  year={2018},
  publisher={American Association for the Advancement of Science}
}

@article{ratnakumar2026predicting,
  title={Predicting sit-to-stand motions with a deep reinforcement learning based controller under idealized exoskeleton assistance},
  author={Ratnakumar, Neethan and Akba{\c{s}}, K{\"u}bra and Jones, Rachel and You, Zihang and Zhou, Xianlian},
  journal={Multibody System Dynamics},
  volume={66},
  number={4},
  pages={835--852},
  year={2026},
  publisher={Springer}
}

@phdthesis{ratnakumar2025optimizing,
  title={Optimizing Hip and Knee Assistance for Walking and Sit-to-Stand Transitions: An Intrinsic Muscle Mechanics Based Predictive Approach},
  author={Ratnakumar, Neethan},
  year={2025},
  school={New Jersey Institute of Technology}
}

@article{yuan2026smat,
  title={{SMAT}: Staged Multi-Agent Training for Co-Adaptive Exoskeleton Control},
  author={Yuan, Yifei and Androwis, Ghaith and Zhou, Xianlian},
  journal={Proceedings of 2026 IEEE/RSJ International Conference on Intelligent Robots and Systems, arXiv preprint arXiv:2603.07618},
  year={2026}
}

@article{brockway1987derivation,
  title={Derivation of formulae used to calculate energy expenditure in man.},
  author={Brockway, JM},
  journal={Human nutrition. Clinical nutrition},
  volume={41},
  number={6},
  pages={463--471},
  year={1987}
}

@article{barati2026end,
  title={End-to-End Policy Learning for Hip Exoskeleton via Reinforcement Learning and Reflex-Based Musculoskeletal Simulation},
  author={Barati, Hossein and Kim, Sangdo and Xuan, Nguyen Thanh and Lee, Jongwon and Park, Young Jin},
  journal={IEEE Robotics and Automation Letters},
  year={2026},
  publisher={IEEE}
}

@article{park2026learning,
  title={Learning hip exoskeleton control policy via predictive neuromusculoskeletal simulation},
  author={Park, Ilseung and Song, Changseob and Kang, Inseung},
  journal={arXiv preprint arXiv:2603.04166},
  year={2026}
}

@article{kang2019effect,
  title={The effect of hip assistance levels on human energetic cost using robotic hip exoskeletons},
  author={Kang, Inseung and Hsu, Hsiang and Young, Aaron},
  journal={IEEE Robotics and Automation Letters},
  volume={4},
  number={2},
  pages={430--437},
  year={2019},
  publisher={IEEE}
}

\clearpage
\onecolumn

\setcounter{section}{0}
\setcounter{table}{0}
\setcounter{figure}{0}
\setcounter{algorithm}{0}
\renewcommand{\thesection}{S\arabic{section}}
\renewcommand{\thetable}{S\arabic{table}}
\renewcommand{\thefigure}{S\arabic{figure}}
\renewcommand{\thealgorithm}{S\arabic{algorithm}}

\renewcommand{\arraystretch}{1.1}
\renewcommand{\topfraction}{0.95}
\renewcommand{\bottomfraction}{0.95}
\renewcommand{\textfraction}{0.05}
\renewcommand{\floatpagefraction}{0.85}

\begin{center}
{\LARGE\bfseries Supplementary Material}\\[6pt]
{\large Staged Multi-Agent Training for Human--Exoskeleton Co-Adaptation:\\
Simulation-Based Learning and Real-World Metabolic and Biomechanical Validation}\\[6pt]
{Yifei Yuan, Jakob Wolf, Ghaith Androwis, and Xianlian Zhou}
\end{center}
\vspace{1.5em}

\section{Participant Characteristics}
Eight healthy adults participated in the study. Subject numbering S1--S8 matches the ordering used in the per-subject tables of the main text. Table~\ref{tab:demographics} lists their sex, age, body mass, and height.

\begin{table}[!ht]
\centering
\caption{Participant characteristics ($n=8$).}
\label{tab:demographics}
\setlength{\tabcolsep}{10pt}
\begin{tabular}{lcccc}
\toprule
Subject & Sex & Age (yr) & Mass (kg) & Height (m) \\
\midrule
S1 & M & 28 & 96.5 & 1.80 \\
S2 & M & 44 & 70.0 & 1.73 \\
S3 & M & 27 & 88.0 & 1.78 \\
S4 & F & 26 & 61.0 & 1.66 \\
S5 & M & 18 & 71.4 & 1.80 \\
S6 & F & 28 & 56.0 & 1.68 \\
S7 & M & 20 & 66.5 & 1.68 \\
S8 & F & 25 & 53.8 & 1.68 \\
\midrule
Mean$\pm$SD & 5M/3F & $27.0\pm7.8$ & $70.4\pm15.0$ & $1.73\pm0.06$ \\
\bottomrule
\end{tabular}
\end{table}

\FloatBarrier
\section{Training Details}
This section collects the implementation details of the four-stage curriculum. The formulation itself is described in full in the Staged Multi-Agent Training section of the main text.

\subsection{Curriculum Schedule}
Algorithm~\ref{alg:multistage} summarizes the training schedule, the policy freezing, and the reward activation across the four stages.

\begin{algorithm}[!ht]
\caption{Multi-stage Curriculum for Human--Exoskeleton Co-adaptation}
\label{alg:multistage}
\begin{algorithmic}[1]
\Require Desired walking speed $v^\star$, PPO hyperparameters $(\Delta t,\gamma)$, training horizons $(T_1,T_2,T_3,T_4)$
\Ensure Trained human policy $\pi_h$ and exoskeleton policy $\pi_e$
\State Initialize human policy $\pi_h$, exoskeleton policy $\pi_e$, and shared value function $V_\psi$
\Statex \textit{Stage 1: Human gait learning}
\For{$t=1$ to $T_1$}
    \State Roll out $\pi_h$ and update $(\pi_h,V_\psi)$ using PPO
\EndFor
\State Store trained human policy $\pi_h^{(1)}$
\Statex \textit{Stage 2: Human adaptation to a passive exoskeleton}
\State Attach the passive exoskeleton (mass/inertia only) and load $\pi_h^{(1)}$
\State Set exoskeleton torque $\tau=\mathbf{0}$
\For{$t=1$ to $T_2$}
    \State Roll out $\pi_h$ and update $(\pi_h,V_\psi)$ using PPO
\EndFor
\State Store adapted human policy $\pi_h^{(2)}$
\Statex \textit{Stage 3: Exoskeleton assistance learning}
\State Re-initialize $\pi_e$ and freeze $\pi_h^{(2)}$
\State Set $\tau_{\max}\leftarrow6$~Nm
\State Configure Stage-3 reward: disable hip imitation terms; enable the hip activation reward and the Stage-3 exoskeleton reward
\For{$t=1$ to $T_3$}
    \State Roll out $(\pi_h,\pi_e)$ and update $(\pi_e,V_\psi)$ using PPO
\EndFor
\State Store trained exoskeleton policy $\pi_e^{(3)}$
\Statex \textit{Stage 4: Human--exoskeleton co-adaptation}
\State Load $\pi_e^{(3)}$ and unfreeze $\pi_h^{(2)}$
\State Set $\tau_{\max}\leftarrow25$~Nm
\State Augment the human observation with current exoskeleton torques $[\hat{u}_r,\hat{u}_l]$, preserving existing weights and initializing the new input weights
\State Configure Stage-4 reward: the Stage-4 exoskeleton reward and the action rate penalty
\For{$t=1$ to $T_4$}
    \State Roll out $(\pi_h,\pi_e)$ and jointly update $(\pi_h,\pi_e,V_\psi)$ using PPO
\EndFor
\end{algorithmic}
\end{algorithm}

\FloatBarrier
\subsection{Reward Configuration}
What changes from stage to stage is the reward configuration. Table~\ref{tab:reward_weights} gives the weight of every reward term in each of the four stages, and the per-joint weights inside the imitation terms $r_{\text{qpos}}$ and $r_{\text{qvel}}$ are listed separately in Table~\ref{tab:imitation_weights}.

\begin{table}[!ht]
\centering
\begin{threeparttable}
\caption{Stage-wise reward configuration and weights.}
\label{tab:reward_weights}
\setlength{\tabcolsep}{12pt}
\begin{tabular}{lcccc}
\toprule
Reward term & Stage 1 & Stage 2 & Stage 3 & Stage 4 \\
\midrule
$r_{\text{fwd}}$        & 0.8   & 0.8   & 1.5  & 1.5 \\
$r_{\text{muscle}}$     & 0.01  & 0.01  & 0.15 & 0.15 \\
$r_{\Delta a}$          & 0.005 & 0.005 & 0.05 & 0.05 \\
$r_{\text{hip-act}}$    & \textemdash & \textemdash & 2.0 & 5.0 \\
$r_{\text{exo}}$        & \textemdash & \textemdash & 4.0 & 4.0 \\
$r_{\Delta\tau}$        & \textemdash & \textemdash & \textemdash & 1.0 \\
Imitation ($r_{\text{qpos/qvel}}$) & 1.0 & 1.0 & 1.0\tnote{*} & 1.0\tnote{*} \\
$r_{\text{constraint}}$ & \textemdash & \textemdash & 0.5 & 0.5 \\
$r_{\text{foot}}$       & \textemdash & \textemdash & 0.3 & 0.3 \\
\bottomrule
\end{tabular}
\begin{tablenotes}[flushleft]
\footnotesize
\item[*] All hip DOFs are set to zero in Stages 3--4 while the other joint imitation terms remain active. The magnitude of the imitation terms is carried by the per-joint weights in Table~\ref{tab:imitation_weights}.
\item \textemdash: term inactive. In this table, $r_{\text{exo}}$ denotes the Stage-3 and Stage-4 exoskeleton rewards defined in the main text.
\end{tablenotes}
\end{threeparttable}
\end{table}

\begin{table}[!ht]
\centering
\begin{threeparttable}
\caption{Per-joint imitation weights $w_i$ for the position ($r_{\text{qpos}}$) and velocity ($r_{\text{qvel}}$) tracking terms. Bilateral DOFs use identical left/right weights and are listed once; mtp denotes the metatarsophalangeal (toe) joint.}
\label{tab:imitation_weights}
\setlength{\tabcolsep}{6pt}
\begin{tabular}{lcccc}
\toprule
 & \multicolumn{2}{c}{$r_{\text{qpos}}$} & \multicolumn{2}{c}{$r_{\text{qvel}}$} \\
\cmidrule(lr){2-3}\cmidrule(lr){4-5}
DOF & Stage 1--2 & Stage 3--4 & Stage 1--2 & Stage 3--4 \\
\midrule
pelvis\_tx            & \textemdash & \textemdash & 0.80 & 0.40 \\
pelvis\_ty            & 0.25 & 0.30 & 0.15 & 0.15 \\
pelvis\_tz            & \textemdash & \textemdash & 0.10 & 0.08 \\
pelvis\_tilt          & 0.20 & 0.40 & 0.10 & 0.15 \\
pelvis\_list          & 0.05 & 0.05 & 0.03 & 0.03 \\
pelvis\_rotation      & 0.05 & 0.05 & 0.03 & 0.03 \\
hip\_flexion\tnote{a}   & 0.20 & 0.00 & 0.10 & 0.00 \\
hip\_adduction\tnote{a} & 0.05 & 0.00 & 0.03 & 0.00 \\
hip\_rotation\tnote{a}  & 0.03 & 0.00 & 0.02 & 0.00 \\
knee\_angle\tnote{a}    & 0.20 & 0.30 & 0.10 & 0.10 \\
ankle\_angle\tnote{a}   & 0.15 & 0.15 & 0.08 & 0.08 \\
mtp\_angle\tnote{a}     & 0.02 & 0.02 & 0.01 & 0.00 \\
\bottomrule
\end{tabular}
\begin{tablenotes}[flushleft]
\footnotesize
\item[a] Bilateral DOF; left and right sides use the same weight.
\item \textemdash: DOF not tracked in that term (pelvis translation along $x$/$z$ is tracked in velocity only).
\end{tablenotes}
\end{threeparttable}
\end{table}

\FloatBarrier
\subsection{PPO Hyperparameters}
The network architecture and the simulation interface stay fixed across stages, and the same PPO settings are reused throughout. Table~\ref{tab:ppo_params} lists these hyperparameters.

\begin{table}[!ht]
\centering
\begin{threeparttable}
\caption{PPO training hyperparameters. Values are shared across stages unless noted otherwise.}
\label{tab:ppo_params}
\setlength{\tabcolsep}{12pt}
\begin{tabular}{lcc}
\toprule
Parameter & S1 & S2--S4 \\
\midrule
Learning rate         & $5{\times}10^{-5}$ & $3{\times}10^{-5}$ \\
PPO clip range        & 0.15   & 0.15 \\
Rollout steps / env   & 2048   & 2048 \\
Minibatch size        & 16384  & 16384 \\
Epochs per update     & 20     & 20 \\
Discount $\gamma$     & 0.99   & 0.99 \\
GAE $\lambda$         & 0.95   & 0.95 \\
Target KL             & 0.01   & 0.01 \\
Max gradient norm     & 0.5    & 0.5 \\
Entropy coefficient   & 0.001  & 0.001 / 0.003\tnote{a} \\
Parallel environments & 32     & 32 \\
\bottomrule
\end{tabular}
\begin{tablenotes}[flushleft]
\footnotesize
\item[a] 0.003 for Stages 3 and 4.
\end{tablenotes}
\end{threeparttable}
\end{table}

\end{document}